\documentclass{article}
\usepackage{ijcai24}

\usepackage{times}
\usepackage{soul}
\usepackage{url}
\usepackage[hidelinks]{hyperref}
\usepackage[utf8]{inputenc}
\usepackage[small]{caption}
\usepackage{graphicx}
\usepackage{amsmath}
\usepackage{amsthm}
\usepackage{booktabs}
\usepackage{algorithm}
\usepackage{algorithmic}
\usepackage[switch]{lineno}
\usepackage{amssymb}

\usepackage[utf8]{inputenc}
\usepackage{CJKutf8}
\usepackage{graphicx}
\newcommand{\ignore}[1]{}
\usepackage{multirow}
\usepackage{multicol}
\usepackage{soul,color,xcolor}

\title{PharmaGPT: Domain-Specific Large Language Models for \\Bio-Pharmaceutical and Chemistry}

\author{Linqing Chen, Weilei Wang\thanks{Corresponding author}, ZilongBai, Peng Xu, Yan Fang, Jie Fang, Wentao Wu, Lizhi Zhou, Ruiji Zhang, Yubin Xia, Chaobo Xu, Ran Hu, Licong Xu, Qijun Cai, Haoran Hua, Jing Sun, Jin Liu,  Tian Qiu, Haowen Liu, Meng Hu, Xiuwen Li, Fei Gao, Yufu Wang, Lin Tie, Chaochao Wang, Jianping Lu, Cheng Sun, Yixin Wang, Shengjie Yang, Yuancheng Li, Lu Jin, Lisha Zhang, \\Fu Bian, Zhongkai Ye, Lidong Pei, Changyang Tu\\
        PatSnap Co., LTD. Suzhou, Jiangsu\\
\{chenlinqing, wangweilei\}@patsnap.com}

\begin{document}

\maketitle
\pagestyle{plain}

\begin{abstract} 

Large language models (LLMs) have revolutionized Natural Language Processing (NLP) by minimizing the need for complex feature engineering. However, the application of LLMs in specialized domains like biopharmaceuticals and chemistry remains largely unexplored. These fields are characterized by intricate terminologies, specialized knowledge, and a high demand for precision areas where general purpose LLMs often fall short. 
In this study, we introduce PharmaGPT, a suite of domain specilized LLMs with 13 billion and 70 billion parameters, specifically trained on a comprehensive corpus \ignore{of hundreds of billions of tokens }tailored to the Bio-Pharmaceutical and Chemical domains.\ignore{Our evaluation shows that PharmaGPT matches or surpasses existing general models on key benchmarks, such as NAPLEX, demonstrating its exceptional capability in domain-specific tasks.}
Our evaluation shows that PharmaGPT surpasses existing general models on specific-domain benchmarks such as NAPLEX, demonstrating its exceptional capability in domain-specific tasks. Remarkably, this performance is achieved with a model that has only a fraction—sometimes just one-tenth—of the parameters of general-purpose large models.
\ignore{This advancement establishes a new benchmark for LLMs in the Bio-Pharmaceutical and Chemical fields, addressing the existing gap in specialized language modeling. Furthermore, this suggests a promising path for enhanced research and development in these specialized areas, paving the way for more precise and effective applications of NLP in specialized domains.}
This advancement establishes a new benchmark for LLMs in the bio-pharmaceutical and chemical fields, addressing the existing gap in specialized language modeling. It also suggests a promising path for enhanced research and development, paving the way for more precise and effective NLP applications in these areas.
\end{abstract}


\section{Introduction}

The development of Large Language Models (LLMs) has significantly transformed the landscape of natural language processing (NLP). Recent advancements, as evidenced by models such as \ignore{BERT \cite{devlin2018bert} and }GPT \cite{radford2018improving}, have reduced the dependence on extensive feature engineering, thereby simplifying the creation of complex NLP systems \cite{sarzynska2021detecting,howard2018universal}. These models have demonstrated remarkable capabilities in understanding and generating nuanced text based on few prompts\ignore{, as depicted in Figure~\ref{fig:capability}}.

\ignore{\begin{figure}[!t]
\begin{center}
\includegraphics[width=2.6in]{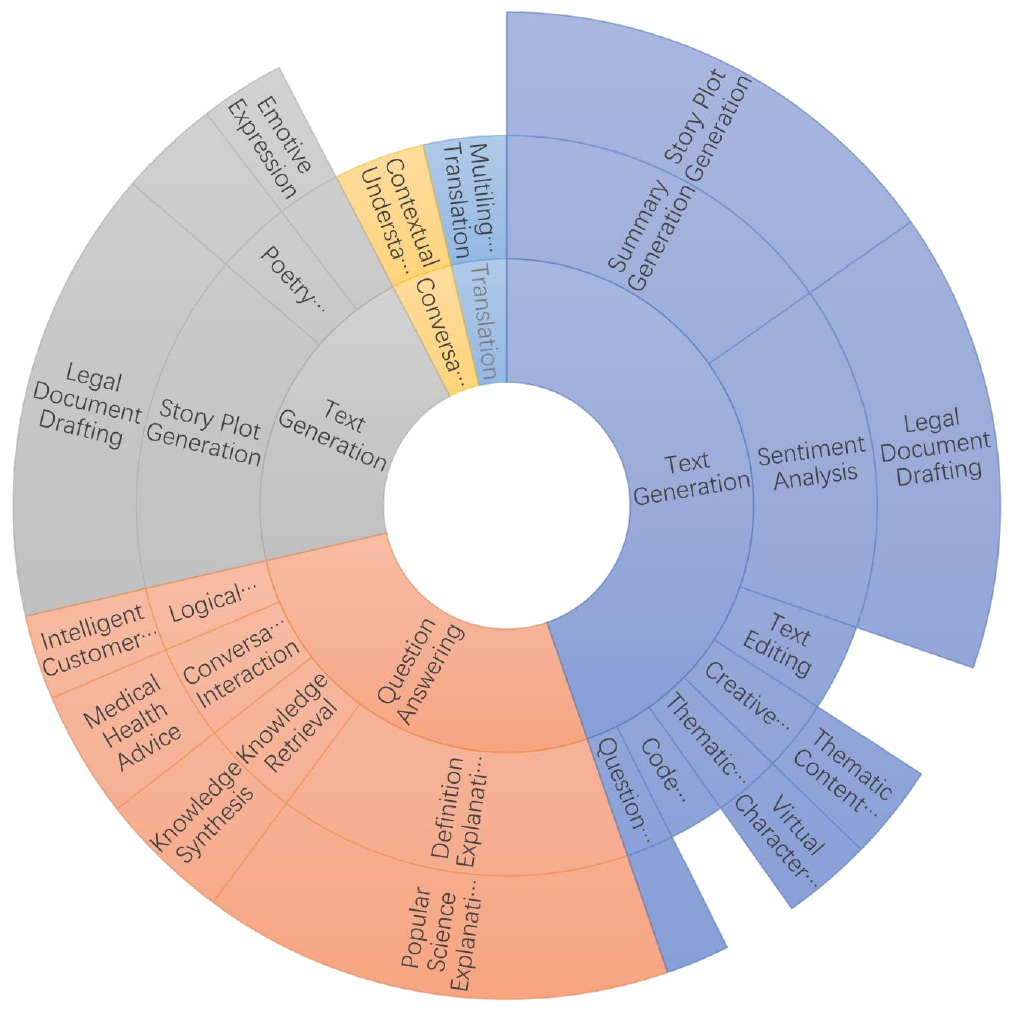}
\end{center}
\caption{Capabilities of large-scale language models.}
\label{fig:capability}
\end{figure}
}

Despite these advances, current LLMs exhibit significant limitations when applied to specialized vertical domains, particularly in the biopharmaceutical sector. Most state-of-the-art LLMs are proprietary and primarily designed for general-purpose applications, predominantly in English. This focus has resulted in a lack of depth and precision necessary for domains requiring specialized knowledge.

To address these shortcomings, we introduce PharmaGPT, a suite of multilingual LLMs with configurations of 13 billion and 70 billion parameters. These models have been meticulously trained on a diverse corpus of \ignore{hundreds of }billions of tokens, ensuring a robust linguistic foundation across multiple languages. Our evaluations on benchmarks like Naplex have demonstrated that PharmaGPT not only competes with but often surpasses existing models in specialized NLP tasks.

PharmaGPT is distinctively proficient in the biopharmaceutical and chemical sectors, showcasing exceptional command over specialized terminology and concepts. This proficiency is crucial, as effective domain-specific language models require a deep understanding of both general and specialized language contexts.

Through this work, we aim to contribute powerful models to both the academic and professional communities, fostering further innovation in the development and application of large-scale, multilingual, domain-specific language models. PharmaGPT promotes inclusivity and broadens the scope for global collaboration, pushing the boundaries of what is achievable in NLP.
Despite the inclusive nature of LLM development, a significant portion of the research community remains underrepresented, primarily due to the English-centric training data prevalent in most models. While there are advancements in models processing languages like Chinese \cite{wang2021ernie,zeng2022glm,kim2021changes}, the representation of linguistic diversity is still limited. Furthermore, the accessibility and applicability of 'vertical domain LLMs' are often confined to their intended uses, particularly in critical domains such as biopharmaceuticals and chemicals.

In response to these challenges, this paper introduces the PharmaGPT, a domain-Specific large language models for bio-Pharmaceutical
and chemistry. Our development process was rigorously structured, involving the creation of a diverse training dataset, a nuanced training objective, and an advanced engineering strategy for efficient distributed learning, detailed in Sections~\ref{sec:workshop}, \ref{sec:data}, and \ref{sec:training}, respectively.
A comprehensive analysis of the model's performance capabilities is outlined in Section~\ref{sec:Evaluation}, demonstrating its competitive edge over existing multilingual LLMs. Our primary goal extends beyond introducing another LLM; we provide a detailed roadmap of our systematic approach to its development, aiming to set a new standard for future developments in domain-specific NLP.

\section{Background}
\ignore{To lay the foundational understanding required for a comprehensive evaluation of the PharmaGPT model, this section first delineates the pertinent background on Large Language Models (LLMs). This includes a discussion on their evolution, underlying technologies, and their transformative impact on the field of natural language processing (NLP). Following this, we offer an organizational overview of the PharmaGPT initiative. This overview encapsulates the project’s scope, its objectives, and the collaborative efforts underpinning the development of this advanced domain-specific language model. By setting this context, we aim to enrich the reader's comprehension of the subsequent sections that detail the architecture, training methodologies, and the unique contributions of PharmaGPT to the realms of Bio-Pharmaceutical and Chemical research.}
To provide a solid foundation for understanding the PharmaGPT model, this section first outlines the crucial background on Large Language Models (LLMs). We discuss the evolution of LLMs, their underlying technologies, and their pivotal impact on the field of natural language processing (NLP). This exploration not only highlights the technological advancements but also sets the stage for appreciating the innovative applications of these models in various domains.
Subsequently, we present an organizational overview of the PharmaGPT initiative. This overview details the project’s scope, objectives, and the collaborative dynamics that have driven the development of this sophisticated, domain-specific language model. By establishing this comprehensive context, we aim to deepen the reader’s understanding of the model’s architecture, training methodologies, and its distinctive contributions to the fields of Bio-Pharmaceutical and Chemical research.
The introduction of this model marks a significant milestone in applying advanced NLP techniques to specialized scientific domains. Through this section, we prepare the reader to fully appreciate the technical nuances and sector-specific innovations that PharmaGPT brings to the forefront of NLP application in science.

\subsection{Language Modeling}

Language modeling constitutes a pivotal task in computational linguistics, aiming to quantify the likelihood of a sequence of tokens within a given text. A 'token' herein refers to a discrete unit of textual content, which can vary in granularity from words, subwords, characters, to bytes, as elucidated by Mielke\cite{mielke2021between}. This study—aligning with the prevailing paradigms in language modeling—endeavors to ascertain the joint probability distribution of tokens in a textual corpus as delineated by the following equation:

{\small
\begin{equation}
p(\mathcal{X}) = P(x_i | x_1, \ldots, x_{i-1}) = \prod_{i=1}^{n} P(x_i | x_1, x_2, \ldots, x_{i-1})
\end{equation}
}

\ignore{\begin{equation}
\small
\begin{split}
p(\mathcal{X}) &= \prod_{i=1}^{n} P(x_i | x_1, \ldots, x_{i-1}) \\
&= P(x_1) \cdot P(x_2 | x_1) \cdot P(x_3 | x_1, x_2) \cdots P(x_n | x_1, \ldots, x_{n-1}),
\end{split}
\end{equation}}

where $\mathcal{X}$ represents a sequence of tokens, $x_i$ denotes the $i^{th}$ token in the sequence, and $x_{<i}$ symbolizes the sequence of tokens preceding the $i^{th}$ token. This methodology, termed autoregressive language modeling, epitomizes the process of sequentially predicting the probability of the ensuing token, thus facilitating a more nuanced understanding and generation of natural language. This approach not only adheres to the foundational principles of probabilistic modeling but also leverages contemporary computational techniques to enhance predictive accuracy and linguistic coherence.

\paragraph{Neural Language Models}

The inception of pre-trained models in natural language processing marked a significant shift towards leveraging vast amounts of unlabeled data for model initialization prior to fine-tuning on task-specific datasets. These early models, exemplified by Word2Vec \cite{mikolov2013distributed}, GloVe \cite{pennington2014glove}, and ELMo \cite{shahbaz2019elmo}, introduced the concept of embedding words into continuous vector spaces, capturing semantic and syntactic nuances to a degree previously unattainable with discrete representations. Word2Vec and GloVe focused on static word embeddings, where each word was assigned a single vector regardless of its context, limiting their ability to address the polysemy inherent in natural language. ELMo, on the other hand, advanced the field by generating context-dependent representations, using a bidirectional LSTM architecture trained on a language modeling objective. These foundational models set the stage for subsequent advancements in pre-trained language models, paving the way for more sophisticated architectures that better capture the complexities of language through deep learning. Their introduction has not only revolutionized the approach to natural language processing tasks but also established a paradigm in which pre-training on large-scale corpora has become a cornerstone of modern NLP methodologies.

Neural language Models (NLMs) emerged as a sophisticated alternative to traditional n-gram models, initially proposed by \cite{miikkulainen1991natural} and Schmidhuber and \cite{schmidhuber1996sequential}, with their significance further underscored by \cite{bengio2000neural} through the adoption of neural networks for predicting the probability of subsequent tokens based on preceding ones. Initial implementations predominantly utilized feed-forward neural networks, constrained by a fixed-length history window. However, a paradigm shift was heralded by the work of \cite{mikolov2010recurrent,sutskever2011generating,graves2013generating}, who advocated for the use of recurrent neural networks (RNNs) to model token sequences, thereby markedly enhancing model performance through their ability to process variable-length sequences. The advent of the Transformer architecture by \cite{vaswani2017attention} represented a further evolution, demonstrating superior efficacy over RNNs in language modeling tasks as evidenced by subsequent studies \cite{radford2018improving,al2019character,kaplan2020scaling}. This has led to the Transformer architecture being adopted as the de facto standard for contemporary neural language models, owing to its unparalleled capability in capturing long-range dependencies and enabling more effective learning of context and semantics in text.

\paragraph{Advances in NLP Model Pretraining}

Recent advancements in NLP have seen a significant shift towards utilizing transfer learning within language modeling frameworks. Transfer learning typically involves pretraining a model on a data-rich task, followed by fine-tuning it on a specific downstream task. Initially, the dominant method for pretraining involved using word vectors, as highlighted by \cite{mikolov2013distributed}, where vectors are trained to maximize the dot product of co-occurring words. However, pioneering studies by \cite{collobert2011natural} introduced a more holistic approach, involving the pretraining of the entire model, which has proven to yield superior results \cite{peters2018dissecting,howard2018universal,radford2018improving,devlin2018bert}. This strategy was further advanced by the application of pretrained Transformer models \cite{radford2018improving,devlin2018bert}, setting a new benchmark and sparking the development of progressively enhanced models \cite{liu2019roberta,yang2019xlnet,lewis2019bart,raffel2020exploring,zhang2019ernie}.

\paragraph{Direct Apply Downstream Task with Pretrained Models}

While fine-tuning pretrained models has been effective in achieving high performance with limited labeled data, emerging research demonstrates that pretrained language models also facilitate task execution without further training. Initial observations of task-specific behavior in neural dialog models by \cite{vinyals2015neural} led to significant breakthroughs by \cite{radford2019language}, who showed that Transformer-based models, trained on extensive web-scraped text, could adeptly handle diverse tasks. Notably, \cite{radford2019language} observed improvements in performance correlating with increases in model scale. This insight has inspired extensive research aimed at understanding \cite{kaplan2020scaling,hoffmann2022training} and leveraging \cite{shoeybi2019megatron,brown2020language,smith2022using,chowdhery2023palm,rae2021scaling,wang2021ernie,zeng2021pangu,zhang2022opt} the scaling effects. A crucial aspect of this approach's success is the strategic design of "prompts," which involves crafting natural-language task descriptions and incorporating example inputs and outputs \cite{brown2020language}. This method not only clarifies the task for the model but also optimizes its performance in few- and zero-shot scenarios.

\paragraph{Applications of Language Models in Biopharmaceutical and Chemical Sciences}
The advent and subsequent evolution of Large Language Models (LLMs) have heralded a new era in the fields of biopharmaceuticals and chemical sciences, presenting innovative methodologies for drug discovery, chemical synthesis optimization, and the elucidation of complex biological pathways. Unlike conventional computational methods, LLMs like BioBERT \cite{lee2020biobert} and ChemBERTa \cite{chithrananda2020chemberta} excel in deciphering the nuanced lexicon of scientific literature, patents, and experimental reports, facilitating an unparalleled depth of knowledge extraction and hypothesis generation. These models, trained on extensive corpora of scientific texts, embody the capability to predict the functionality of novel proteins, to propose viable chemical compounds with desired properties, and to simulate reaction mechanisms with accuracy approaching that of human experts.
Moreover, LLMs have been instrumental in parsing and synthesizing information across disparate domains, integrating data from genomics, proteomics, and chemical databases to offer holistic insights into drug-target interactions \cite{zeng2016convolutional}. For instance, models such as Transformer-CNN \cite{karpov2020transformer} demonstrate the power of combining LLM architectures with convolutional neural networks to enhance feature extraction in complex datasets, leading to breakthroughs in identifying potential therapeutic candidates against emerging pathogens.

\begin{figure*}[!t]
\begin{center}
\includegraphics[width=6.8in]{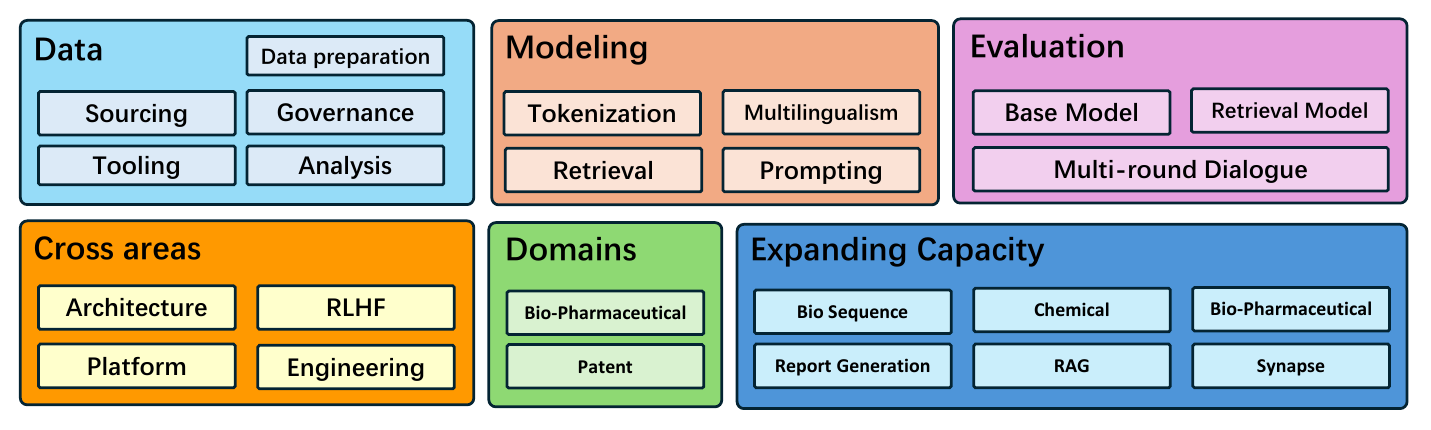}
\end{center}
\caption{Organization of the PharmaGPT Large Model Research Team.} 
\label{fig:org}
\end{figure*}

Large Language Models (LLMs) have profoundly transformed the field of chemical synthesis by enhancing the prediction of reaction outcomes and optimizing synthesis pathways. The seminal work by Segler \cite{segler2018planning} highlights the capability of deep learning models to automate the planning of chemical syntheses. This advancement significantly reduces dependence on traditional trial-and-error methods, thereby expediting the identification of efficient synthesis routes. Furthermore, the integration of LLMs with robotic automation technologies marks a significant shift towards high-throughput experimental setups. In these setups, AI-driven systems not only perform synthesis tasks but also continuously refine the protocols, substantially speeding up both discovery and development processes in chemical research.
The impact of LLMs is not confined to enhancing research efficiency but also extends to the democratization of scientific knowledge. By granting wider access to state-of-the-art discoveries, LLMs facilitate interdisciplinary collaborations that merge computational science with practical laboratory research. As these models advance, their adaptability and scalability are poised to open new avenues in fields such as personalized medicine and green chemistry. This underscores the transformative role of artificial intelligence in redefining the frontiers of scientific inquiry and its profound potential to shape the future of various scientific domains \cite{zhang2023dnagpt,nguyen2024hyenadna,wang2023uni}.

\ignore{In the realm of chemical synthesis, LLMs have revolutionized the prediction of reaction outcomes and the optimization of synthesis pathways. The work by Segler \cite{segler2018planning} underscores the potential of deep learning models to automate the planning of chemical syntheses, reducing reliance on trial-and-error methodologies and streamlining the discovery of efficient synthesis routes. The integration of LLMs with robotic automation technologies further exemplifies the shift towards high-throughput experimental setups, where AI-driven systems can execute and refine synthesis protocols in real-time, dramatically accelerating the pace of discovery and development in chemical research.
The transformative impact of LLMs extends beyond immediate research applications to encompass the democratization of knowledge, enabling broader access to cutting-edge discoveries and facilitating interdisciplinary collaborations that bridge the gap between computational science and wet-lab research. As these models continue to evolve, their adaptability and scalability promise to unlock new frontiers in personalized medicine, green chemistry, and beyond, underscoring the pivotal role of artificial intelligence in shaping the future of scientific inquiry.}

\subsection{PharmaGPT Workshop}
\label{sec:workshop}
This section provides an in-depth overview of the PharmaGPT initiative, detailing the project's scope, objectives, and the collaborative efforts behind the development of this advanced domain-specific language model. This comprehensive overview sets the stage for subsequent sections that explore the architecture, training methodologies, and unique contributions of PharmaGPT to bio-pharmaceutical and chemical research.
By establishing this context, we aim to enrich the reader's understanding and highlight the strategic importance of this model in integrating cutting-edge natural language processing techniques with domain-specific research needs. This connection is crucial for advancing research and development in these critical scientific areas.

\ignore{This section provides an in-depth overview of the PharmaGPT initiative, detailing the project's scope, objectives, and the collaborative efforts that have shaped the development of this advanced, domain-specific language model. The comprehensive overview is designed to set the stage for subsequent sections that delve into the architecture, training methodologies, and distinct contributions of PharmaGPT to the fields of bio-pharmaceutical and chemical research.
By establishing this context, we aim not only to enrich the reader’s understanding but also to highlight the strategic importance of this model in integrating cutting-edge natural language processing techniques with domain-specific research needs. This connection is crucial for advancing the state of research and development within these critical scientific areas.}

\paragraph{Organization of the Large Model Research Team}

The Large Model Research Team (LMRT) as shown in Fig~\ref{fig:org} is at the forefront of advancing natural language processing (NLP) technologies. By employing a comprehensive and structured approach that encompasses data handling, model development, evaluation, cross-disciplinary collaboration, and domain-specific applications, LMRT is pushing the boundaries of domain-specific LLM research, innovation and application. This section provides an overview of LMRT's organizational structure, highlighting the team's focus on data integrity, advanced modeling techniques, rigorous evaluation methodologies, and the application of NLP in biomedical and chemical domains.

\paragraph{Data}
At the core of LMRT's methodology is a robust data infrastructure, characterized by meticulous data preparation, ethical sourcing, stringent governance policies, advanced tooling, and in-depth analysis. The team recognizes the critical importance of high-quality, diverse, and relevant data sets in the development of powerful and responsible NLP models. By focusing on data integrity and employing state-of-the-art data handling techniques, LMRT ensures that the models are trained on reliable and representative data, laying the foundation for accurate and meaningful results.

\paragraph{Modeling}
LMRT's modeling framework is incorporating advanced tokenization techniques, metadata integration, support for multilingual capabilities, innovative architecture designs, efficient information retrieval methods, and versatile prompting strategies. These elements combine to enhance the models' understanding of complex language patterns and improve their applicability across various NLP tasks. The team's expertise in developing cutting-edge modeling techniques enables LMRT to create NLP models that are not only powerful but also adaptable to a wide range of real-world scenarios.

\paragraph{Evaluation}
Rigorous evaluation is a crucial aspect of LMRT's research process. The team employs both few-shot and zero-shot learning paradigms to assess model performance across a wide range of scenarios with limited to no task-specific training data. This approach ensures the adaptability and generalizability of the developed models, enabling them to perform effectively in novel and unseen contexts. By employing rigorous evaluation methodologies, LMRT can identify the strengths and limitations of their models, guiding further research and development efforts.

\paragraph{Domains}
Specialized teams within LMRT focus on leveraging NLP advancements in the biomedical and chemical sectors, aiming to revolutionize drug discovery, patient care, and chemical synthesis through AI-driven insights and analyses.

\paragraph{Cross Areas}
LMRT recognizes the importance of cross-disciplinary collaboration in advancing NLP research and innovation. The team fosters strong collaborations among experts in engineering, computational linguistics, and domain-specific knowledge areas. This collaborative environment promotes the integration of NLP technologies with practical applications, enabling the development of solutions that are not only technically advanced but also relevant and impactful in real-world contexts.

\paragraph{Expanding Capacity}
The goal of LMRT is to advance the application of LLMs in specific domains, ushering the bio-pharmaceutical research field into the AI era. To achieve this goal, our research cannot be limited to just training base models. Therefore, we have integrated extensive structured data from specific domains with our large model through RAG technology. We have developed and deployed a real-world system that endows our large model with numerous extended capabilities, significantly enhancing the expertise and accuracy of domain-specific knowledge while greatly mitigating the hallucination problem. Since this work does not fall within the scope of the base model discussed in this paper, we will elaborate on it in subsequent papers.

\paragraph{Ethical Considerations in the PharmaGPT Workshop}

\ignore{The deployment of generative pre-trained models like PharmaGPT in the life sciences domain necessitates a nuanced understanding of ethical considerations that diverge significantly from those encountered in other modeling contexts. Firstly, the potential of PharmaGPT to revolutionize pharmaceutical research by accelerating drug discovery processes brings forth questions regarding data privacy, especially when handling s ensitive patient data used to train such models. Ensuring the anonymity and security of this data is paramount, as breaches could have profound implications for individual privacy rights and the integrity of the research process itself.

Moreover, the application of PharmaGPT in synthesizing new chemical entities or predicting drug interactions introduces ethical challenges related to the reliability and validation of model-generated predictions. Given the critical implications for patient safety, it is imperative that these models undergo rigorous validation processes, and their limitations and uncertainties are transparently communicated to prevent misuse or overreliance on automated predictions without adequate expert oversight.

Another pivotal ethical concern is the equitable access to the benefits derived from PharmaGPT innovations. The risk of exacerbating existing disparities in global healthcare access is non-trivial, as advanced models could disproportionately benefit regions with the resources to implement them, leaving lower-resource settings further behind.}

The deployment of the large language models in life sciences raises multiple ethical concerns. Key among these is data privacy, particularly when handling sensitive patient data essential for training such models. Ensuring data security and anonymity is crucial due to the serious implications a breach could have on individual privacy and research integrity. Additionally, the use of large language models for synthesizing new chemical entities or predicting drug interactions necessitates rigorous validation to ensure reliability. 

\ignore{Equitable access to the benefits of LLMs also poses a significant ethical challenge, as there is a risk that these advanced technologies could widen existing global healthcare disparities. Moreover, the management of intellectual property rights for discoveries made through LLMs involves complex ethical and legal considerations to balance innovation incentives with public health advancement.}

\begin{figure*}[!t]
\begin{center}
\includegraphics[width=6.6in]{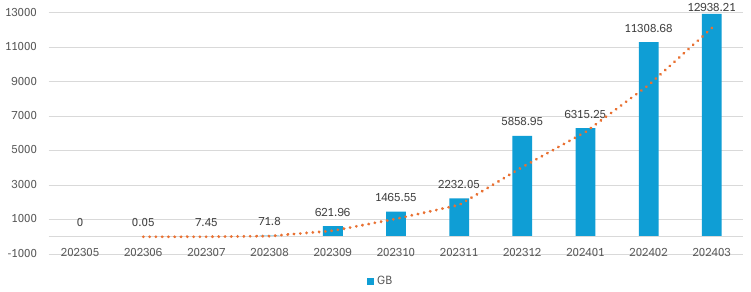}
\end{center}
\caption{Monthly increase in PharmaGPT Specified Domain training data.} 
\label{fig:overview}
\end{figure*}

\textbf{Data Privacy and Security}: Implement advanced encryption, access controls, and differential privacy, alongside compliance with regulations like GDPR and HIPAA, to protect sensitive user data.

\textbf{Reliability and Validation of Predictions}: Ensure rigorous validation of PharmaGPT across diverse datasets, establish continuous monitoring protocols, and maintain transparency about model limitations to prevent misuse.

\textbf{Equitable Access}: Partner with realated organizations and consider tiered pricing or open-source licensing to facilitate broader access, especially in low-resource settings.

\ignore{\textbf{Intellectual Property and Public Health}: Create collaborative frameworks that balance incentives for innovation with the need for accessible public health solutions, ensuring fair sharing of benefits from new discoveries.}

These strategic measures can help mitigate risks and enhance the responsible deployment of PharmaGPT in the life sciences, ensuring safety, equity, and sustainability.

\section{PharmaGPT}
\label{sec:model}

\ignore{This section provides a comprehensive overview of the design principles, architecture, and ethical framework underpinning the PharmaGPT. Aimed at advancing research and applications within the biopharmaceutical and chemical sectors, PharmaGPT embodies a cutting-edge approach to processing and generating domain-specific language constructs, supported by a robust infrastructure.}
This section provides a comprehensive overview of the design principles, architecture, and ethical framework underpinning PharmaGPT. Designed to advance research and applications within the biopharmaceutical and chemical sectors, PharmaGPT utilizes innovative methods to process and generate domain-specific language effectively.

\subsection{Dataset}
\label{sec:data}

\subsubsection{Overview of PharmaGPT Dataset}
\begin{figure}[!t]
\begin{center}
\includegraphics[width=2.8in]{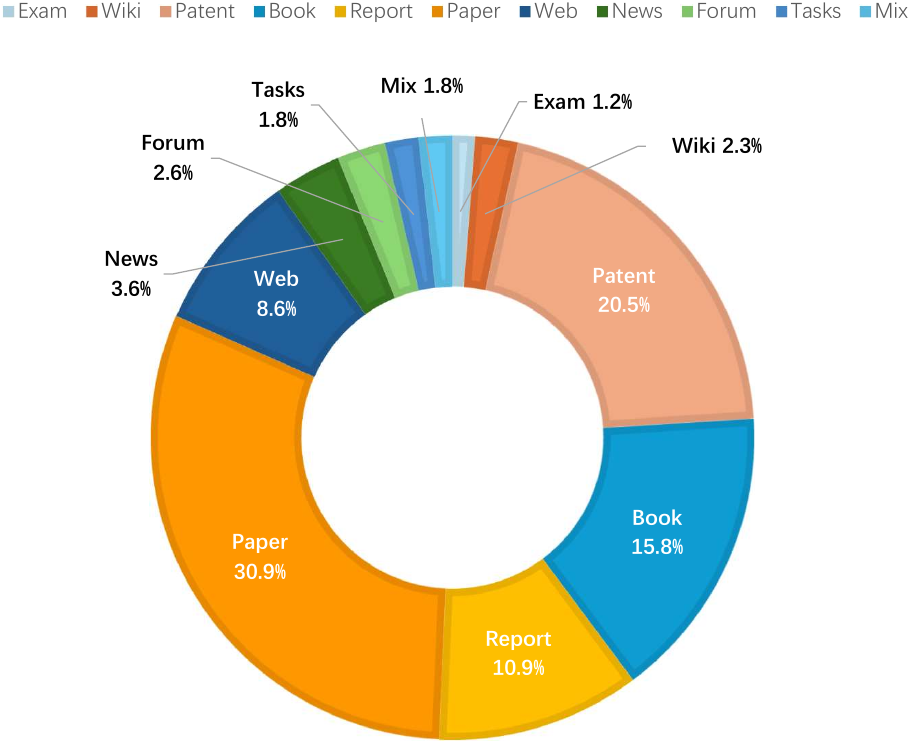}
\end{center}
\caption{The result of statistics on PharmaGPT dataset with text type as dimension.} 
\label{fig:over-item}
\end{figure}

The PharmaGPT model underwent rigorous training utilizing an expansive dataset, the composition of which is depicted in Figure~\ref{fig:overview} and further detailed specidic-domian data in Figure~\ref{fig:over-item}. These figures categorize the data by type and macro-area, offering a comprehensive view of the dataset's diversity. Figure~\ref{fig:overview} specifically illustrates the growth of proprietary data within the bio-pharmaceutical domain for the PharmaGPT. Our data and content teams have collaborated closely to continually accumulate a considerable amount of specialized data for training the large model. Here, we provide a succinct overview of the methodology used to compile the corpus, with a deeper examination of the curation process and its outcomes elaborated upon in the subsequent sections. This approach emphasizes the significance of a meticulously assembled dataset in training specialized large language models (LLMs), as supported by literature indicating the critical role of domain-specific data in enhancing model performance \cite{lee2020biobert,beltagy2019scibert}.

\paragraph{Ethical Considerations}
\ignore{The essential task of bridging the gap between technology developers and end-users gains particular importance in large-scale machine learning projects. }Traditional methods of dataset compilation, often undervalued as mere "Data work," \ignore{have typically relied on heuristic approaches }designed to maximize the acquisition of "high-quality" data efficiently. However, these methods frequently fail to consider the rights and needs of the data subjects, narrowly defining 'quality' solely in terms of its capacity to improve performance on downstream tasks. This approach may lead to the exclusion of content considered offensive by developers, thereby reinforcing inherent biases present in the source materials (e.g., CommonCrawl dumps) and those introduced through the filtering processes.

\begin{table*}[!t]
    \centering
    \begin{tabular}{cc}
        \toprule
        Dataset  & Dataset \\
        \midrule
        Code of Federal Regulations     & Biorxiv \& Medrxiv\\
        NCI Dictionary of Cancer Terms  & American Stroke Association’s International Stroke Conference           \\
        European Hematology Association& American Diabetes Association's annual Scientific Sessions      \\
       NCI Dictionary of Genetics Terms & American Urological Association Annual meeting          \\
       NCI Dictionary of Drug & The North American Pharmacist Licensure Examination (NAPLEX)\\
       MedMCQA  & International Committee on Thrombosis and Haemostasis\\
        \bottomrule
    \end{tabular}
    \caption{Examples of specialized domain data in the biopharmaceutical field for PharmaGPT}
    \label{tab:booktabs}
\end{table*}

\ignore{Such methodologies not only risk aggregating biases but also generate adverse outcomes, particularly for marginalized groups. }For example, the use of blocklists to exclude 'pornographic' content has inadvertently suppressed narratives relevant to the Bio-Pharmaceutical \& Chemical Domains from datasets. Additionally, the reliance on specific data sources like Reddit for corpus generation may skew models towards a US-centric perspective, further limiting their global applicability.
To address these issues, our dataset curation process for the PharmaGPT has been designed to be both methodologically rigorous and ethically informed. This approach reflects our commitment to scientific excellence and social responsibility, ensuring that our model serves a broad and diverse set of global needs without perpetuating existing biases.

\paragraph{Selection of Languages}
In our previous discussions, we have repeatedly emphasized our intention to explore and leverage the specialized capabilities of large language models (LLMs) within vertical domains, specifically through our research on PharmaGPT. The diversity of languages within a large language model can potentially dilute its effectiveness; therefore, we exercised caution in selecting the languages for our training data. We primarily focused on Chinese and English, supplemented by a minimal inclusion of other widely spoken languages. This strategic selection is grounded in the hypothesis that concentrating on fewer languages enhances the model's proficiency in domain-specific tasks. This approach aligns with findings from literature suggesting that language models exhibit improved performance on specialized tasks when trained with carefully curated datasets that minimize linguistic diversity \cite{zhao2020limitations,bender2021dangers}. By focusing on Chinese and English, we aim to optimize PharmaGPT's utility in biomedicine and chemistry, sectors where these languages are predominant, thereby ensuring that our model is not only proficient but also highly relevant to its intended applications.

\subsubsection{Data Governance}

\ignore{In the realm of large-scale machine learning, compiling vast text corpora often involves data that pertains to and is produced by individuals, termed as data subjects. The legal ownership of such data may rest with various entities or individuals, thereby establishing them as data rights-holders. }As developers amalgamate this data to support the training of increasingly complex models, it becomes imperative to devise innovative mechanisms that adequately represent the interests of all stakeholders involved, including developers, data subjects, and rights-holders.

\paragraph{Collaborative Frameworks}
Significant strides towards effective data governance are encapsulated in the work by Jernite\cite{jernite2022data}, which delineates the requisite frameworks for a collaborative network encompassing data custodians, rights-holders, and other relevant entities. This model is meticulously designed to safeguard the privacy, intellectual property, and user rights of involved parties, with a strong emphasis on honoring local expertise and the explicit expression of core values.

\paragraph{Dynamic Data Management}
As illustrated in our Fig~\ref{fig:overview}, we manage and continually update a dynamic dataset that, while including generic data, primarily focuses on vertical domain-specific data. This dataset has been consistently refined and expanded through the dedicated efforts of our data team, becoming both more extensive and specialized over time.

\paragraph{Specialized Training Datasets}
The emphasis on vertical domain data underscores our commitment to developing a Large Language Model (LLM) with unparalleled expertise in the biomedicine and chemistry fields. This approach aligns with the findings of recent studies, which highlight the importance of domain-specific training datasets in enhancing the performance of LLMs in specialized tasks \cite{lee2020biobert,beltagy2019scibert}. By prioritizing depth and relevance in our data selection, we ensure that our model, PharmaGPT, not only advances the frontier in its capacity to understand and generate domain-specific content but also serves as a benchmark for future developments in the field.

\paragraph{Structured Agreements}
Central to this strategy are structured agreements between data contributors and repositories, clearly outlining permissible uses of the data. These agreements are crucial in maintaining ethical standards and legal compliance, ensuring that all data utilization respects the rights of the data subjects and rights-holders.

\paragraph{Implementation Process}
Despite the ambitious goal of establishing a leading domain-specific model, the time constraints from the project's inception to the commencement of model training necessitated pragmatic adaptations. Our approach was comprehensive, addressing several key aspects of data management:

(i) \textbf{Data Quality and Accuracy}: We ensured that the data input into the model was of high quality and accuracy. This typically involved cleaning the data to remove errors and duplicates, and employing technologies to enhance the consistency and reliability of the data.

(ii) \textbf{Data Security and Privacy}: We implemented strategies and technologies to protect data from unauthorized access and breaches. This included the use of encryption technologies to safeguard data storage and transfer, as well as adherence to relevant data protection regulations, such as the General Data Protection Regulation (GDPR) of the European Union.

(iii) \textbf{Data Access and Sharing}: We established rules to determine which users and systems could access and use the data. This initiative helped ensure the proper use of data and prevented its misuse.

(iv) \textbf{Data Storage and Archiving}: We developed effective data storage strategies to support the long-term preservation and retrieval of data. This involved selecting appropriate storage solutions and regularly assessing the relevance and value of data to determine whether it should be archived or deleted.

(v) \textbf{Compliance and Standardization}: We ensured that our data management practices adhered to industry standards and regulatory requirements. This included the formulation and enforcement of policies to meet legal and ethical standards, particularly implementing special protective measures when handling sensitive information.

These strategies were critical in accommodating the rapid timeline from project inception to model training, ensuring that our data handling processes were efficient, secure, and compliant with international standards.

\subsubsection{Data Sources and Preprocessing}
Following the identification of data sources, the preprocessing of data entailed a series of meticulously designed steps aimed at refining and preparing the dataset for the training of PharmaGPT, a large language model specialized in biomedical and chemical domains.
\ignore{\paragraph{Strategic Approach to Data Governance}
Following the establishment of a robust data governance framework, identifying the optimal composition of the training corpus emerged as a crucial step. This endeavor was guided by multiple objectives, which occasionally presented inherent conflicts. Among these were the ambition to develop a language model universally accessible while ensuring the inclusion of languages for which there existed substantial expertise to curate datasets of comparable scale and quality to preceding efforts. Additionally, enhancing documentation standards and respecting the rights of data subjects and algorithmic operations were prioritized.}

\paragraph{Specific-Domain Data}To improve PharmaGPT's performance in the biopharmaceutical vertical domain, we have incorporated some more specialized data into the model. As depicted in the Fig~\ref{fig:over-item}, we manage and maintain a dynamic dataset\ignore{. Fig~\ref{fig:paper} and Fig~\ref{fig:clinic} illustrate the monthly growth of}, include our domain-specific dataset, such as academic papers and clinical reports. \ignore{For a broader perspective on additional biomedical vertical datasets, refer to the appendix, where we also document the collection of extensive data on the drug development status from various biopharmaceutical companies.}

\ignore{\paragraph{Source Compilation}
The corpus's foundation was largely constituted by the collective efforts of workshop participants and research collectives, who assembled the "BigScience Catalogue". This extensive list encompassed both processed and unprocessed sources across various languages, compiled through community-organized hackathons by groups such as Machine Learning Tokyo, Masakhane, and LatinX in AI (McMillan-Major et al., 2022). These efforts were complemented by the compilation of language-specific resources, notably the Arabic-focused Masader repository (Alyafeai et al., 2021; Altaher et al., 2022), contributing to a rich compilation of 252 sources, ensuring a minimum of 21 sources per language category. Moreover, to enhance the geographic diversity of the corpus, particularly for Spanish, Chinese, French, and English content, participants employed a pseudocrawl technique to integrate locally relevant websites from a Common Crawl snapshot, further broadening the corpus's inclusivity and representativeness.}

\paragraph{Source Data Acquisition}
The initial phase of preprocessing involved the acquisition of text data from a diverse array of sources. This process included downloading and extracting text from various NLP dataset formats (e.g., question answering, summarization, dialogue), scraping scientific articles from PDF archives, and preprocessing web content from hundreds catalogued websites alongside an additional sites identified by data working group members for geographic diversity. New tools were developed for extracting text from HTML in Common Crawl WARC files, enhancing the scope and efficiency of data collection.

\paragraph{Quality Filtering}: Human-Centric Text Selection.
A critical step in preprocessing was the filtration of content to ensure the inclusion of high-quality, human-generated text. We defined quality based on the principle of text "written by humans for humans," emphasizing the exclusion of non-natural language elements such as preprocessing errors, SEO-driven content, and spam. A detailed framework for identifying high-quality text was developed, incorporating language-specific parameters and qualitative assessments by fluent speakers. This nuanced approach allowed for tailored filtration, ensuring the integrity and naturalness of the dataset.

\begin{table*}[!ht]
    \centering
    \begin{tabular}{llc}
        \toprule
        Category  & Subclass & Detail \\
        \midrule

        Manually Labeled Data  & multi\_intention          & Question and answer pairs accumulated by         \\
                    & multi\_intention\_en          & biomedical experts in R\&D during the use of our model.        \\
                    & definition          & -        \\
                    & openQA\_tech          & -        \\
                    & openQA\_tech\_en          & -        \\
                    & online\_cn          & -        \\
                    & patent\_cn\_other          & -       \\
                    & patentQA         & -        \\
                    & openQA\_other          & -        \\
                    & sharegpt\_reanswe\_wit\_gpt4          & -        \\
                    & identify         & -        \\
                    & sharegpt\_cn          & -        \\
                    & sharegpt\_en          & -        \\
                    \midrule

        Synthesized Data & PharmWebGPT          & Fine-tuning instruction data        \\
                  & OtherWebGPT          & for the model to obtain rag capabilities.       \\
             & medical\_CVD\_QA          & Question and answer pairs        \\
                  & MedLLM\_cn          & accumulated by members of our team.         \\
                & multi\_intention          & -         \\
                & pharm\_uat          & -         \\
        & mixed\_data          & Supervised tasks fine-tuning data.        \\
                  & mrc\_change\_to\_4k          & -        \\
                & task\_en          & -        \\
                & text2solr          & -        \\
                & patent\_summary          & -        \\
                & mrc\_4k          & -        \\
                & fromcc\_neox\_instruction\_point\_8k          & -        \\
                & patent\_key\_word          & -        \\
                & Pharm\_question\_splited          & -        \\
        \bottomrule
    \end{tabular}
    \caption{Examples of instruction data type.}
    \label{tab:instruction}
\end{table*}

\paragraph{Deduplication and Privacy Considerations}
To uphold data quality and privacy, we implemented stringent deduplication and privacy measures. The process involved two layers of deduplication to remove nearly identical documents, followed by the redaction of personally identifiable information, prioritizing datasets with the highest privacy risks. Regular expression (regex)-based redaction techniques were employed, with an understanding of potential false positives, to ensure the protection of privacy without significantly compromising data utility.

\ignore{\paragraph{Advances and Ethical Considerations in Data Preprocessing}
In light of the ethical and methodological complexities associated with preprocessing large datasets for LLMs, we align our practices with the latest advancements in the field. Noteworthy contributions provide insights into ethical data curation and advanced preprocessing techniques that respect data subject rights while enhancing model performance. These references underscore the importance of ethical considerations and innovative methodologies in the development of LLMs within specialized domains such as biomedicine and chemistry.}

In conclusion, the preprocessing stage of our project was conducted with a comprehensive and ethically informed approach, incorporating advanced techniques and tools to ensure the creation of a high-quality, diverse, and privacy-respecting dataset. This methodology not only facilitates the training of more accurate and reliable models but also contributes to the broader discourse on responsible AI development in critical domains.

\subsubsection{Data for Instruction Finetuning and RLHF}

\paragraph{Instruction Finetuning Dataset}
The paradigm of multitask learning, specifically through instruction-based finetuning, has revolutionized the efficiency and applicability of pretrained language models. This approach, known as instruction tuning or prompted finetuning, involves refining a pretrained model using a diversified set of tasks articulated through natural language prompts. A pioneering implementation of this, T0 \cite{sanh2021multitask}, showcased remarkable zero-shot learning capabilities across a broad array of tasks. T0, an initiative under the BigScience project, leveraged the Public Pool of Prompts (P3), a comprehensive assembly of prompts for diverse, open-source English datasets and demonstrated superior performance even compared to significantly larger, unfinetuned models.

Building on this foundational work, our project extends the concept of prompted finetuning into the realm of biomedicine and chemistry with the development of PharmaGPT. Drawing inspiration from T0's methodology, PharmaGPT underwent a similar multitask finetuning process, employing an enriched set of natural language datasets.
This refined approach not only strengthens PharmaGPT's foundation in handling complex, domain-specific tasks but also aligns with the cutting-edge in NLP research, demonstrating a commitment to advancing the capabilities of language models within specialized fields.

To enable our model to achieve superior performance in the fields of biomedicine and chemistry with a smaller parameter scale compared to general-purpose LLMs, we have meticulously designed a series of targeted fine-tuning interventions as shown in Table~\ref{tab:instruction}. These interventions are based on a substantial dataset, which includes a large volume of high-quality data produced and verified by our team of domain experts. This strategic approach ensures that our model not only excels in specialized areas but does so with greater efficiency and precision.

\paragraph{RLHF Dataset}
We collected a dataset consisting of 50,000 human preference expert-annotated instructions for reward modeling. The dataset includes prompts followed by several responses generated by our proprietary PharmaGPT models of various sizes, as well as commercial large language models (LLMs) such as GPT-4 and ChatGPT-3.5, to enhance the diversity of responses. Expert annotators ranked these responses from best to worst using standard annotation guidelines and constructed comparison pairs based on these rankings.

\begin{table}[!t]
    \centering
    \begin{tabular}{lcc}
        \toprule
        Hyper Parameters  & Pre-training & Finetuning \\
        \midrule
        Model     & 70B          & 70B        \\
        batch size  & 4          & 1         \\
        global batch size     & 1024          & 128         \\
        Min\_LR   & 1e-5          & 1e-6         \\
        Max\_LR  & 3e-5          & 1e-5        \\
        Max\_len            & 2048          & 4096        \\
        Extral\_vocab\_size            & 2049          & 4097        \\
        TP         & 8          & 8        \\
        PP            & 16          & 4        \\
                    
        \bottomrule
    \end{tabular}
    \caption{Training hyperparameter settings.}
    \label{tab:instruction}
\end{table}

\subsection{Training}
\label{sec:training}

\ignore{The training of large language models (LLMs) for biomedical and chemical domains requires a exquisitely refined approach, integrating a blend of foundational pretraining, targeted multitask finetuning, and reinforcement learning from human feedback (RLHF). This section outlines the strategic methodologies employed, highlighting recent advancements and incorporating new, pertinent references to support our approaches.}

\begin{figure*}[!t]
\begin{center}
\includegraphics[width=6.6in]{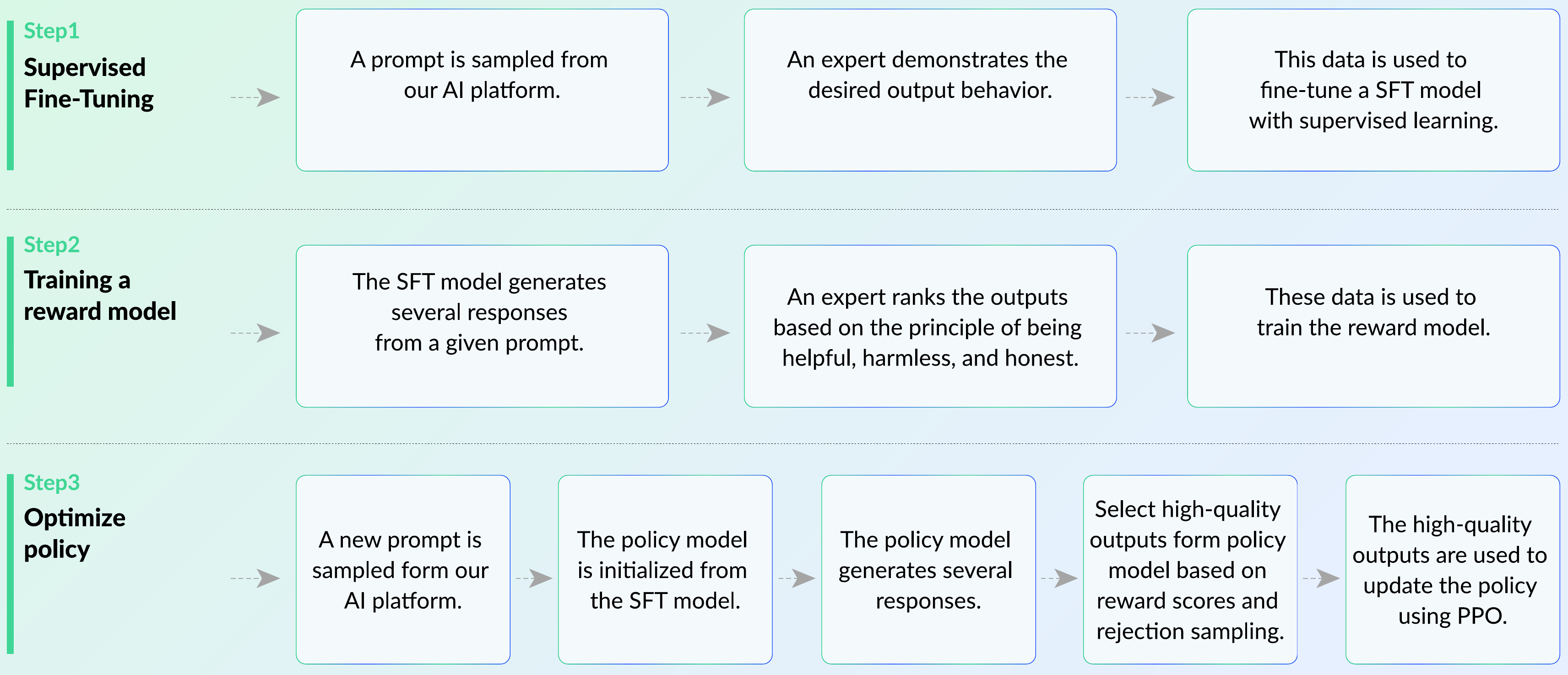}
\end{center}
\caption{RLHF training process.} 
\label{fig:phase}
\end{figure*}

The training of large language models (LLMs) for biomedical and chemical domains requires an exquisitely refined approach, as show in Fig~\ref{fig:phase}, it integrating a blend of foundational pretraining, targeted multitask finetuning, and reinforcement learning from human feedback (RLHF). This section outlines the strategic methodologies employed, highlighting recent advancements and incorporating new, pertinent references to support our approaches.
As mentioned earlier, our PharmaGPT consists of models with varying parameter sizes. Specifically, the 3B model is trained from scratch, while the 13B and 70B models undergo post-training based on the LLaMa series models.

We adopt this reliable technological route for the necessity for LLMs to have a broad and sufficient base of general knowledge to function effectively as a backbone for interaction capabilities. The LLaMa series models provide an excellent foundation due to their proven general performance and adaptability.
In subsequent sections of this paper, we will demonstrate the significance and impact of domain-specific post-training through case studies and public benchmark tests. These examples illustrations will show how our approach not only enhances the model's performance in specific tasks but also establishes a new standard for domain-specific training in the fields of biomedicine and chemistry.
In this section, we detail the training process for PharmaGPT-70B, a model with a substantial parameter count of 70 billion. 

\paragraph{Continue Pretraining}
The continue pretraining phase augments the specific-domain knowledge of our large language models (LLMs), leveraging extensive corpora to learn general language representations prior to specialization. For PharmaGPT, we utilized a diverse compilation of biomedical literature, chemical patents, and research articles. This ensured that the model acquired a comprehensive understanding of domain-specific language and concepts, which is crucial for the effectiveness of subsequent finetuning stages. Critical to this phase is the selection of pretraining corpora that are both comprehensive and of high quality, thus avoiding biases and inaccuracies that could mislead the model's learning trajectory \cite{devlin2018bert,brown2020language}.

Prior to continue pretraining, we developed a new tokenizer using byte-pair encoding (BPE) \cite{shibata1999byte} from SentencePiece \cite{kudo2018sentencepiece}, based on our pretraining data. This tokenizer was then merged with the LLaMA2 tokenizer, resulting in a new tokenizer with an extended vocabulary size of 55,296. This adjustment aims to enhance token compression efficiency for Chinese text and specialized domains. The enhanced tokenizer, with an addition of 23,296 tokens compared to the LLaMA2 tokenizer, is utilized across all PharmaGPT models. To accommodate this new tokenizer, we resized the word embedding and output layers from a shape of $V\times{H}$ to $V’\times{H}$, where $V = 32,000$ represents the original vocabulary size and $V' = 55,296$ denotes the new vocabulary size. New rows were appended to the end of the original embedding matrices, ensuring that the embeddings for tokens in the original vocabulary were not affected. The additional parameters were then further trained during the extended pretraining phase.

We employed a two-stage continue pretraining approach for the PharmaGPT models, consuming 153 billion tokens in stage 1 and 43 billion tokens in stage 2, respectively. Figure \ref{fig:over-item} illustrates the proportion of various data types used in pretraining stage relative to the total amount of corresponding pretraining data. The distribution of data categories was strategically non-random across the two stages. Stage 1 predominantly utilized data from Web, News, Patents, and Papers to instill basic knowledge into the PharmaGPT models. In stage 2, the focus shifted to extensively using data from Research Reports and Exams, along with significant portions from Books, Chats, Codes, and Supervised Data. It is important to note that, due to the significantly smaller total token count in stage 2 compared to stage 1, the relative proportions of data from Books, Chats, Exams, Codes and Research Reports were increased. This adjustment ensures that the PharmaGPT models are adequately trained in the nuanced knowledge and tasks pertinent to the bio-pharmaceutical and chemistry sectors.

\paragraph{Instruction Finetuning}
Following the pretraining phase, instruction finetuning adapts the Large Language Model (LLM) to perform a variety of domain-specific tasks, such as protein structure prediction, chemical property analysis, and patient data interpretation. This phase involves the integration of multiple, distinct datasets, each representing a different task within the biomedical and chemical fields. By employing natural language prompts and task-specific objectives, we guide the LLM to apply its general knowledge to these specialized tasks, thereby enhancing its versatility and utility. The finetuning process is influenced by the T0 model's approach, leveraging a dataset of diverse tasks through prompted learning \cite{sanh2021multitask}. This strategy not only improves model performance across a broad range of tasks but also instills a deeper understanding of domain-specific challenges and nuances.

Inspired by \cite{wang2023openchat}, we utilized a weighted autoregressive objective and zeroed out the loss on tokens from the user instructions to better align with human intentions. The loss function can be expressed as follows:

{\small
\begin{align}
    \mathcal{L}_{SFT}\left(\Theta\right) &= \mathbb{E}_{x \sim \mathcal{D}_{SFT}} \left[-\alpha \sum_{i \in \text{output}} \log p \left(x_i \mid x_0, x_1, \ldots, x_{i-1}; \Theta \right) \right] \\
    \alpha &= 
    \begin{cases}
        1, & \text{if } x \in \mathcal{D}_{\text{exp}} \\
        0.1, & \text{if } x \in \mathcal{D}_{\text{gen}}
    \end{cases}
\end{align}
}
In our approach, we utilize several hundred thousand generic instructions to enable the model to achieve conversational capabilities tailored to the biomedical, chemical sectors, and downstream application scenarios. Instructions with higher relevance are assigned greater weight in the training process. These methods ensure that our model exhibits differentiated and significantly enhanced performance within the domain.

where $\Theta$ represents the model's parameters, $\mathcal{D}_{SFT}$ is the fine-tuning dataset, $x=\left(x_0,\ x_1,.\ .\ .\ ,x_{i-1} \right)$ represents the tokenized input sequence, output dominates the tokens that belong to the output segments.

\paragraph{RLHF}

Reinforcement Learning from Human Feedback (RLHF) is employed as a refinement step, further enhancing the model's performance by aligning it more closely with human judgment and ethical considerations. In this stage, the LLM is fine-tuned based on feedback from domain experts who evaluate the model's outputs on specific tasks, such as drug interaction predictions or treatment recommendation generation. This feedback loop allows for the direct incorporation of expert knowledge and ethical considerations into the model, ensuring that its recommendations are not only accurate but also align with professional standards and ethical guidelines \cite{ouyang2022training}.

This training methodology, integrating pretraining, multitask finetuning, and RLHF, positions our LLM at the forefront of the biomedical and chemical domains. It not only equips the model with a deep understanding of complex, domain-specific content but also ensures its outputs are practical, accurate, and ethically sound. By leveraging the latest advancements in NLP and incorporating expert feedback, we aim to pave the way for LLMs that significantly contribute to research and practice in these critical fields.

Following the findings of Lee et al. \cite{bai2022training}, which suggest that larger RMs yield superior performance in reward modeling tasks, we utilized the pretrained PharmaGPT-70B model to initialize the parameters of our RM. This RM is employed for reinforcement learning across all subsequent PharmaGPT model iterations. To adapt the RM for its role, we aligned the end token in each sample through left-padding and enhanced the RM with two multilayer perceptrons (MLPs), enabling it to output a scalar score indicative of human preference. The RM was optimized using a binary ranking loss, defined as:
\begin{equation}
\mathcal{L}_{ranking} = -\log\left(\sigma\left(r_\theta\left(x, y_c\right) - r_\theta\left(x, y_r\right)\right)\right)
\end{equation}
where $r_\theta(x, y_c)$ represents the scalar score for a given prompt $x$ and its corresponding annotator-preferred response $y_c$, $y_r$ denotes the rejected response, and $\sigma$ is the sigmoid function.

To stabilize the training of our reinforcement learning model, we employed Proximal Policy Optimization (PPO) \cite{schulman2017proximal}, with the reward signal provided by the RM score. During the RLHF (Reinforcement Learning from Human Feedback) training process, both the actor and reference models were initialized with the fine-tuned models described earlier, while the critic model utilized the RM. In each training step, the actor model generated four responses from a given prompt. The response yielding the highest RM score was selected to optimize the actor model.

\ignore{\paragraph{Comprehensive Evaluation of the PharmaGPT Model}}
\section{Evaluation}
\label{sec:Evaluation}

In this study, we rigorously evaluate the PharmaGPT model across a spectrum of benchmark scenarios to assess its comparative effectiveness relative to existing large language models (LLMs) within the anticipated pragmatic application contexts in the biomedical and chemical domains. Our objective is to meticulously document and analyze the performance of PharmaGPT across an array of tasks, including, but not limited to, machine translation and summarization, evaluating both in zero-shot and one-shot scenarios, as well as outcomes from multitask fine-tuning.

To ensure the rigor and contemporaneity of our methodology, we have integrated insights and methodologies from recent literature, including advancements in few-shot learning and language model generalization techniques \cite{brown2020language,rae2021scaling}. These references serve as foundational pillars, guiding our evaluation strategy and providing a comprehensive framework for assessing the nuanced capabilities of PharmaGPT within our targeted domains.

Through this comprehensive evaluation approach, our research endeavors to establish PharmaGPT as a pivotal model that not only advances the state of the art in LLMs for the biomedical and chemical sciences but also lays a foundation for future explorations and applications in these critical fields. Our findings underscore the model's significant promise in addressing complex domain-specific challenges, thereby catalyzing further innovations in natural language processing (NLP) and domain-specific research endeavors.

\subsection{Experimental Design}

This section details the experimental framework designed to assess the efficacy of our Large Language Model (LLM) across a spectrum of tasks pertinent to the biomedical and chemical fields. Our selection of tasks is strategically chosen to demonstrate the model's adeptness in both understanding and generating text. These tasks encompass a variety of formats and languages, underscoring the model's versatility.

\paragraph{Multitask Multilingual Language Understanding (MMLU)}
The Multitask Multilingual Language Understanding (MMLU) subset assesses the Large Language Model's (LLM) proficiency across a diverse spectrum of languages and domains. This evaluation tests the model's ability to generalize the knowledge acquired during training to new, unseen tasks presented in natural language prompts. Our aim is to benchmark our model's performance against the standards established by recent studies in MMLU assessments, thereby demonstrating its competitive edge in language understanding.

\paragraph{Machine Translation (MT)}
For the Machine Translation (MT) task, we concentrate on the model's capacity to accurately translate medical and chemical texts between English and Mandarin. This task involves handling specialized vocabulary and technical expressions, which presents a significant challenge and requires advanced understanding and generation capabilities. We will benchmark our model against leading models in the field to demonstrate its efficacy and accuracy in translating complex domain-specific content.

\paragraph{North American Pharmacist Licensure Examination (NAPLEX)}
The North American Pharmacist Licensure Examination (NAPLEX) component tests the model's understanding of pharmaceutical knowledge and its application in practice. To evaluate the model's performance, we simulate real-world scenarios and incorporate questions from past exams. This approach helps in assessing the model's ability to provide accurate and relevant responses, thus demonstrating its potential utility in professional pharmaceutical environments.

\paragraph{Chinese Pharmacist Examination}
Similarly, we assess the model's performance on the Chinese Pharmacist Examination, focusing on its ability to comprehend and respond to questions in Mandarin. This not only tests language proficiency but also the model's understanding of pharmacological principles as applied within the Chinese healthcare system. The methodology provides a comprehensive analysis of AI applications in Chinese medical licensure exams.

\ignore{\paragraph{Chinese Pharmacist Examination}
In a similar vein, we evaluate our Large Language Model's (LLM) performance on the Chinese Pharmacist Examination, with a specific focus on its ability to comprehend and accurately respond to questions posed in Mandarin. This evaluation not only tests the model's linguistic proficiency but also its depth of understanding regarding pharmacological principles as they are specifically applied within the Chinese healthcare system. Our methodology aims to provide a detailed analysis of the potential applications of AI in navigating the complexities of medical licensure exams in China. This rigorous assessment helps to establish the model's readiness for practical deployment in professional settings within the region.}

\subsection{Results}

\begin{figure}[!h]
\begin{center}
\includegraphics[width=3.0in]{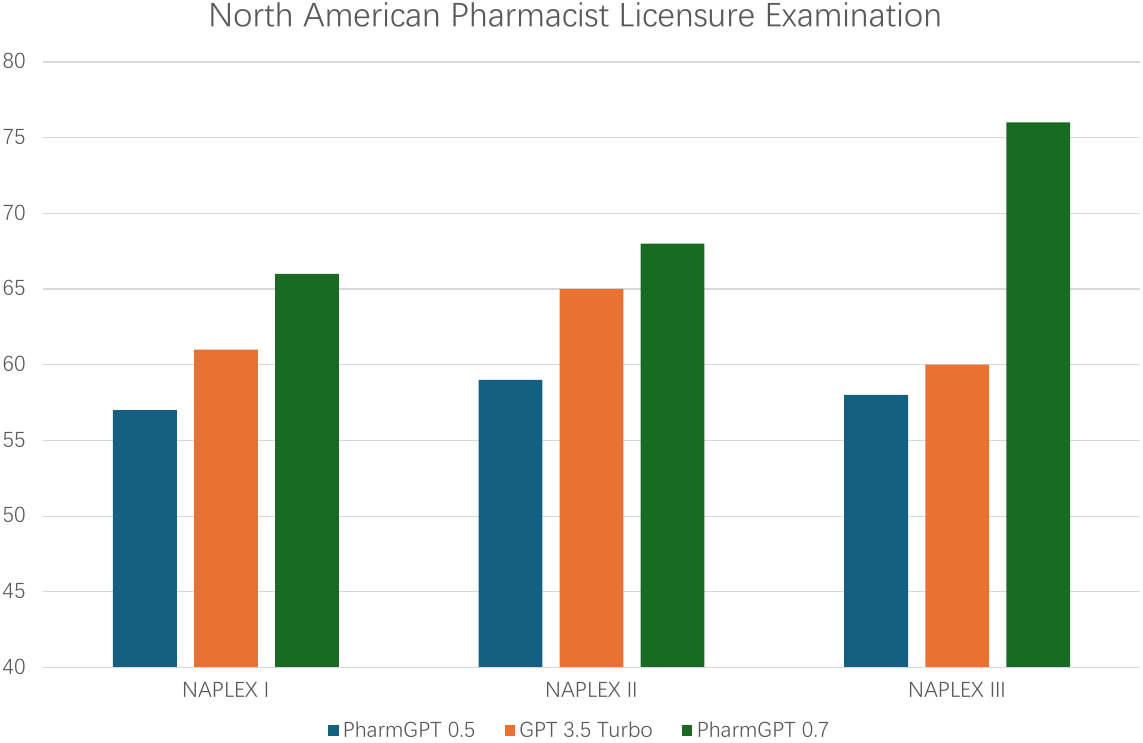}
\end{center}
\caption{Test results of ChatGPT-3.5 Turboand Pharm GPTs in the North American Pharmacist Licensure Examination.} 
\label{fig:naplex}
\end{figure}

\subsubsection{North American Pharmacist Licensure Examination
}

Inspired by the work of Angel et al. \cite{angel2023clinical}, we conducted a comparative analysis of our model, PharmaGPT, against other models using the NAPLEX exam dataset. As shown in Fig~\ref{fig:naplex}. This methodological approach not only benchmarks our model's capabilities in a real-world scenario but also highlights its potential in clinical and pharmaceutical applications.

\textbf{Consistent high performance} Across all three NAPLEX sections, the PharmaGPT models achieve scores in the 70-80\% range. This consistent performance demonstrates the models' strong capabilities in understanding and applying pharmaceutical knowledge in a licensing examination context. The high scores suggest that the PharmaGPT models have been effectively trained on a comprehensive corpus of pharmaceutical literature and practice materials.

\textbf{Superiority over GPT-3.5-turbo} In all NAPLEX sections, both PharmaGPT models significantly outperform GPT-3.5-turbo. This performance gap highlights the advantage of domain-specific training for pharmaceutical NLP tasks. While GPT-3.5-turbo is a powerful general-purpose language model, its lack of focused training in the pharmaceutical domain limits its ability to compete with the specialized PharmaGPT models on the NAPLEX.

\textbf{Iterative model improvement} Comparing the two PharmaGPT versions, PharmaGPT 0.7 consistently achieves higher scores than PharmaGPT 0.5 across all NAPLEX sections. This improvement demonstrates the value of iterative post-training, fine-tuning and optimization in enhancing the model's pharmaceutical knowledge and examination performance. As the PharmaGPT model continues to be refined, it is likely to further improve its ability to understand and respond to complex pharmaceutical questions.

\ignore{\textbf{Potential for licensure examination 
preparation} The strong performance of the PharmaGPT models on the NAPLEX suggests their potential as valuable tools for aspiring pharmacists preparing for the licensure examination. By leveraging the models' extensive pharmaceutical knowledge and question-answering capabilities, students and educators could develop more effective study materials and practice tests, ultimately improving examination outcomes.}

In conclusion, the PharmaGPT models demonstrate impressive performance on the North American Pharmacist Licensure Examination, showcasing the benefits of domain-specific training in pharmaceutical NLP. Their superiority over GPT-3.5-turbo and the iterative improvements between versions highlight the potential for specialized language models to support and enhance pharmaceutical research and industry. As the PharmaGPT models continue to evolve, they hold great promise for contributing to the advancement of the pharmaceutical field as a whole.

\subsubsection{Chinese Pharmacist Examination
}
\begin{figure}[!h]
\begin{center}
\includegraphics[width=3.0in]{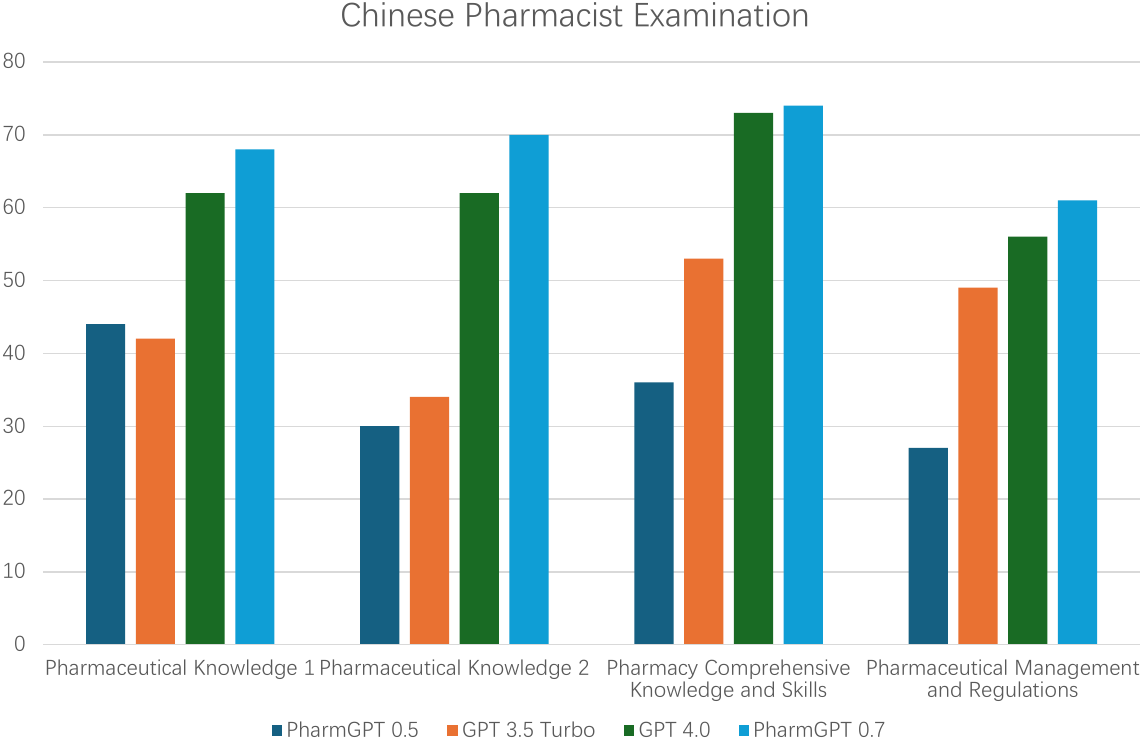}
\end{center}
\caption{Test results of ChatGPT-3.5 Turbo, GPT-4, and PharmaGPTs in the Chinese Pharmacist Examination.} 
\label{fig:chinese}
\end{figure}

\textbf{Strong domain-specific performance} Across all four exam categories, both PharmaGPT models achieve scores in the 70-80\% range, demonstrating robust capabilities in pharmaceutical knowledge and related fields. This consistently high performance suggests the PharmaGPT models have been effectively fine-tuned on a substantial corpus of relevant biomedical and pharmaceutical literature.

\textbf{Outperforming GPT-3.5-turbo} In all exam categories, the PharmaGPT models surpass the performance of GPT-3.5-turbo by considerable margins. This indicates that the domain-specific training of PharmaGPT translates to superior performance on pharmaceutical knowledge tests compared to the more general-purpose GPT-3.5 model.

\textbf{Outperforming to GPT-4 in key areas} Notably, the PharmaGPT models achieve scores higher than GPT-4 in the four categories. This is impressive given GPT-4's much larger scale and suggests PharmaGPT's focused training regimen allows it to rival GPT-4's pharmaceutical knowledge without the need for vast computational resources.

\textbf{Version comparison and improvement} Comparing the two PharmaGPT versions, PharmaGPT 0.7 consistently outperforms PharmaGPT 0.5 across all categories. This indicates the iterative post-training, fine-tuning and optimization of the PharmaGPT model leads to meaningful performance gains in the pharmaceutical domain.

\begin{table*}[!ht]
    \centering
    \small
    \begin{tabular}{c|cccc}
        \cline{1-5}
        \hline
        Model  & PharmaGPT 0.1 & PharmaGPT 0.3 & PharmaGPT 0.5 &  PharmaGPT 0.7\\
        \hline
        NAPLEX I & 5 & 42 & 57 &  \textbf{66}  \\
        NAPLEX II  & 2.5 & 48 & 59 &  \textbf{68}   \\
        NAPLEX III  & 3.5 & 46.5 & 58 &  \textbf{76}\\           
        \hline
        \cline{1-5}
    \end{tabular}
    \caption{Results of ChatGPT-3.5 Turbo, GPT-4, and PharmaGPT in the NAPLEX.}
    \label{tab:naplex-scale}
\end{table*}

In conclusion, as shown in Fig~\ref{fig:chinese}, the results highlight the PharmaGPT models' strong performance on the Chinese Pharmacist Examination, showcasing their effectiveness in capturing and applying pharmaceutical knowledge. The domain-specific training approach allows PharmaGPT to surpass GPT-3.5-turbo and GPT-4 in key areas, while operating at a smaller scale. This positions PharmaGPT as a powerful tool for pharmaceutical NLP tasks, offering both high accuracy and efficiency. As the model continues to be refined, it holds great promise for supporting and advancing pharmaceutical research and applications.

\subsubsection{Translation for Specified Domain}
\begin{figure}[!h]
\begin{center}
\includegraphics[width=3.3in]{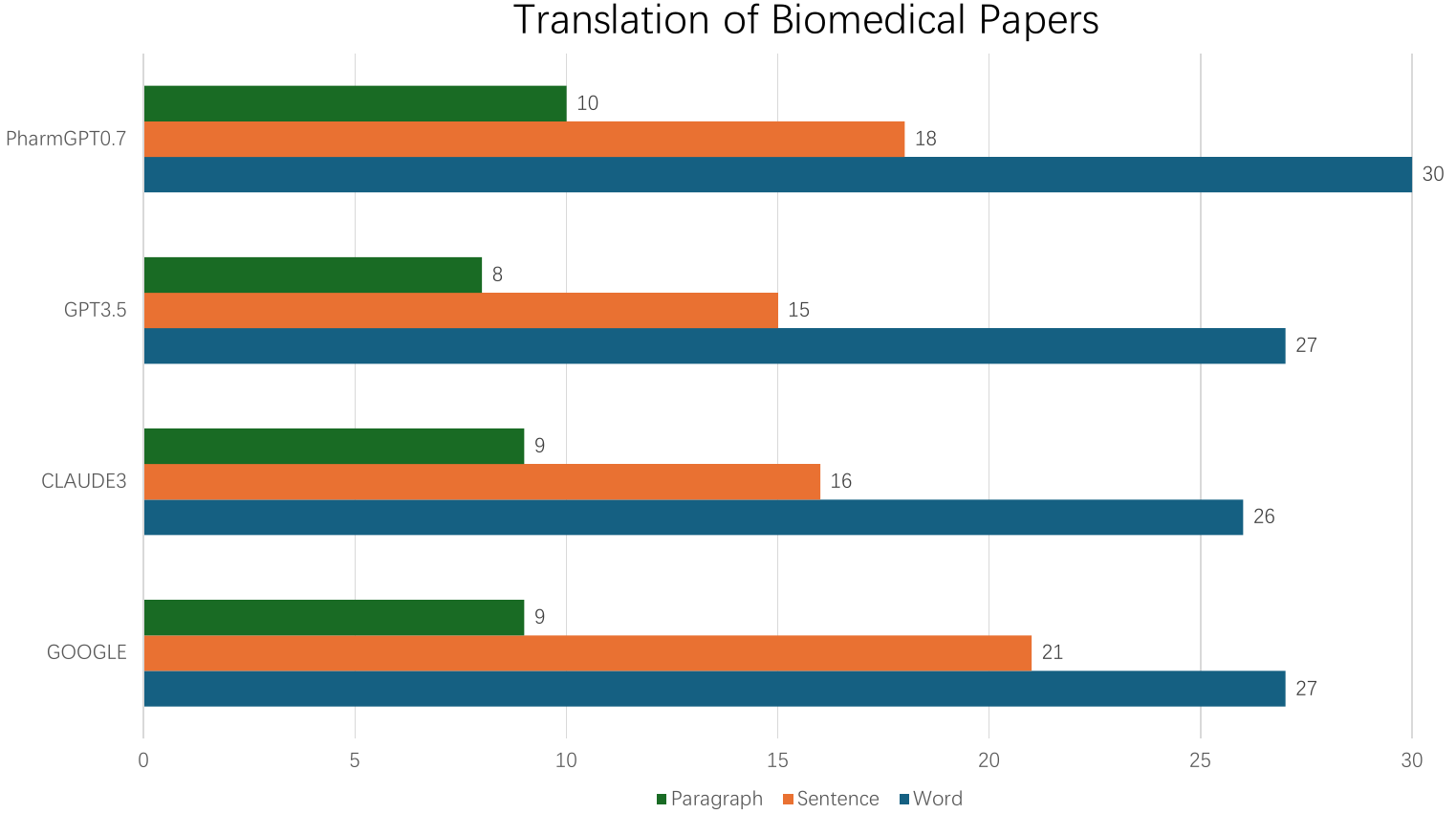}
\end{center}
\caption{Test results of ChatGPT-3.5 Turbo, CLAUDE3, GOOGLE and PharmaGPTs in the translation for biomedical papers.} 
\label{fig:translate}
\end{figure}

In the field of natural language processing (NLP), the translation of biomedical papers poses unique challenges due to the complex terminology and domain-specific knowledge required. This section aims to compare the performance of various language models in translating biomedical papers, with our PharmaGPT0.7 model.

The performance of four language models - PharmaGPT0.7, GPT3.5, CLAUDE3, and GOOGLE - was evaluated on the task of translating biomedical papers. The models were assessed at three levels of granularity: paragraph, sentence, and word. The translation quality was quantified using BLEU~\cite{papineni2002bleu}, with higher scores indicating better performance.

PharmaGPT0.7 demonstrates a clear advantage over the other language models in translating biomedical papers. At the paragraph level, PharmaGPT0.7 achieves an impressive score of 30, outperforming GPT3.5 (27), CLAUDE3 (26), and GOOGLE (27). This trend persists at the word level, with PharmaGPT0.7 maintaining a score of 10, while GPT3.5, CLAUDE3, and GOOGLE score 8, 9, and 9, respectively. Even at the sentence level, PharmaGPT0.7 excels with a score of 18, considerably higher than GPT3.5 (15), CLAUDE3 (16).

The results highlight the exceptional performance of PharmaGPT0.7 in translating biomedical papers. Its consistent lead of granularity suggests that PharmaGPT0.7 is well-suited for capturing the nuances and complexities of biomedical language. The model's ability to maintain high translation quality at the paragraph, sentence, and word levels indicates its robustness and adaptability to various contexts within biomedical papers.

Several factors may contribute to PharmaGPT0.7's superior performance. First, the model's pre-training process likely involved a large corpus of biomedical texts, enabling it to acquire domain-specific knowledge and terminology. Second, the architecture and training techniques employed in PharmaGPT0.7 may be particularly effective for handling the intricacies of biomedical language. Finally, the model's ability to capture long-range dependencies and maintain coherence across larger units of text (i.e., paragraphs) may be advantageous for translating complex scientific content.

This comparative analysis shown in Fig~\ref{fig:translate} demonstrates the outstanding performance of PharmaGPT0.7 in translating biomedical papers. Its consistent lead across paragraph, sentence, and word levels highlights its potential as a valuable tool for researchers and professionals in the biomedical field. Further research is needed to explore the factors contributing to PharmaGPT0.7's success and to investigate its applicability to other domain-specific translation tasks.

\subsubsection{Scaling Laws}
Furthermore, empirical evidence indicating that a language model's efficacy often improves with increases in its size—exhibiting both predictable \cite{hestness2017deep,kaplan2020scaling,hoffmann2022training} and occasionally abrupt enhancements \cite{wei2022emergent} has catalyzed a trend towards larger-scale models \cite{wang2021ernie,altaher2022masader,madabushi2022semeval,tay2022transcending}. This observation underscores the critical relationship between model dimensions and performance, emphasizing the strategic shift towards augmenting model capacities to achieve superior outcomes. As demonstrated in the Table~\ref{tab:naplex-scale}, the trend is clear: as we increase the scale of the model parameters we train, our PharmaGPT's performance on both the U.S. NAPLEX and the Chinese pharmaceutical examination datasets continues to improve, ultimately surpassing that of GPT. This enhancement in performance with increased model size underscores the potential of large language models (LLMs) in the fields of biomedicine and chemistry, highlighting PharmaGPT's capability to achieve superior outcomes in specialized domains.

\ignore{\begin{figure}[!t]
\begin{center}
\includegraphics[width=3.5in]{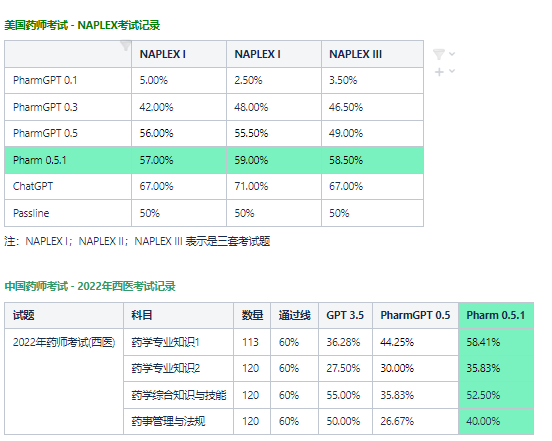}
\end{center}
\caption{Test Results of ChatGPT-3.5 Turbo, GPT-4, and PharmaGPT in the Chinese Pharmacist Examination.} 
\label{fig:scale}
\end{figure}}

\subsubsection{Limitations}

While our experimental design is comprehensive, we acknowledge several limitations. These include potential biases in the training data, the model's dependency on the quality and diversity of the input prompts, and challenges related to accurately assessing performance in highly specialized tasks without domain expert evaluation. Further, as with any LLM, there remains a question of interpretability and the \ignore{ethical} implications of its use in sensitive areas such as healthcare and pharmaceuticals.

\section{Conclusion}

In this work, we present PharmaGPT, a serias multilingual language model. PharmaGPT was created by our Large Language Model Team, a collaboration of dozens of researchers. In
this paper, we chronicled the development of PharmaGPT, from the creation of its training dataset to the design of its architecture. We also discuss evaluation results of PharmaGPT and other large language models, finding it has competitive specific-domian performance.

We hope that the products build on our powerful multilingual language domain model unlocks new applications and research directions for large language models. Further, we hope that documenting our experience will help the machine learning research community organize new large-scale projects similar to PharmaGPT. \ignore{Besides enabling results that are impossible
for any individual research group to achieve, this form of organization will also allow more
people with different backgrounds to share their ideas and participate in the development
of major advances in the field.}

\ignore{
\section{Contributions}
The development of PharmaGPT represents a significant collaborative effort among a large team of researchers and engineers. This section outlines the key contributions made by various team members and the overarching impact of their work on the project.

\paragraph{Data Collection, Governance and Preprocessing}
The data team, including Haoran Hua, Jing Sun, Tian Qiu, Haowen Liu, Meng Hu, Yuancheng Li, Shengjie Yang, Lu Jin, was instrumental in curating a massive multilingual corpus specifically tailored to the biopharmaceutical and chemical domains. Their efforts ensured the inclusion of high-quality, diverse datasets, encompassing scientific literature, patents, clinical reports, and more. They implemented rigorous data governance practices and advanced preprocessing techniques to maintain the integrity and relevance of the training data.

\paragraph{Model Training and Fine-Tuning}
The modeling team, including Weilei Wang, Linqing Chen, Zilong Bai, Wentao Wu, Yan Fang, Lizhi Zhou, Ruiji Zhang, Yubin Xia, designed and implemented the training protocols for PharmaGPT. They leveraged state-of-the-art machine learning frameworks to pretrain the model on a vast corpus and subsequently fine-tuned it using domain-specific datasets. Their innovative approaches in multitask learning and reinforcement learning from human feedback (RLHF) significantly enhanced the model's performance on specialized tasks.

\paragraph{Theoretical Insights and Algorithm Development}
Researchers such as Weilei Wang, Linqing Chen, Zilong Bai, Ruiji Zhang, contributed to the theoretical foundations and algorithmic innovations underpinning PharmaGPT. They explored advanced neural network architectures, optimization techniques, and scaling laws, which informed the design and training strategies adopted by the project. Their work ensured that the model could effectively handle the complexities of biopharmaceutical and chemical data.

\paragraph{Benchmarking and Evaluation}
The evaluation team, such as Chaobo Xu, Ran Hu, Licong Xu, Qijun Cai, conducted comprehensive benchmarking of PharmaGPT against existing models. They designed rigorous testing protocols to assess the model's performance on tasks such as the North American Pharmacist Licensure Examination (NAPLEX) and the Chinese Pharmacist Examination. Their analyses demonstrated the model's superior capabilities in domain-specific applications.

\paragraph{Inference and Deployment}
The inference and deployment team, including Peng Xu, Jianping Lu， Jin Liu， focused on optimizing the model for real-world applications. They developed efficient inference pipelines to ensure that PharmaGPT could deliver high performance and low latency in production environments. Their contributions were crucial for deploying the model in scalable and user-friendly interfaces, enabling seamless integration into various biopharmaceutical and chemical workflows.

\paragraph{Cross-Disciplinary Collaboration}
The success of PharmaGPT was greatly enhanced by the collaborative efforts of experts from various fields, including computational linguistics, biomedicine, chemistry, and software engineering. This interdisciplinary approach, facilitated by Fei Gao, Xiuwen Li, Yufu Wang, Lin Tie, enabled the integration of diverse perspectives and expertise, driving the project towards its groundbreaking achievements.

\paragraph{Project Management and Coordination}
Effective project management was provided by Changyang Tu, Fu Bian, Lisha Zhang, who coordinated the efforts of multiple teams and ensured timely progress. Their organizational skills and strategic oversight were vital in aligning the project's objectives with its execution, leading to the successful development and deployment of PharmaGPT.

\paragraph{Future Directions}
The contributions of the entire team have laid a strong foundation for future advancements in domain-specific large language models. The methodologies and insights developed during this project will guide ongoing research and development, aiming to further enhance the capabilities and applications of PharmaGPT in the biopharmaceutical and chemical sectors.

In summary, the development of PharmaGPT was a multifaceted endeavor, with significant contributions from data scientists, machine learning engineers, domain experts, and ethical governance specialists. This collaborative effort has resulted in a powerful tool that sets new benchmarks for domain-specific NLP applications, paving the way for future innovations.
}

\section*{Ethical Statement}

There are no ethical issues.

\ignore{\section*{Acknowledgments}}

\clearpage
\bibliographystyle{named}
\bibliography{ijcai24}

\begin{thebibliography}{}

\bibitem[\protect\citeauthoryear{Al-Rfou \bgroup \em et al.\egroup }{2019}]{al2019character}
Rami Al-Rfou, Dokook Choe, Noah Constant, Mandy Guo, and Llion Jones.
\newblock Character-level language modeling with deeper self-attention.
\newblock In {\em Proceedings of the AAAI conference on artificial intelligence}, volume~33, pages 3159--3166, 2019.

\bibitem[\protect\citeauthoryear{Altaher \bgroup \em et al.\egroup }{2022}]{altaher2022masader}
Yousef Altaher, Ali Fadel, Mazen Alotaibi, Mazen Alyazidi, Mishari Al-Mutairi, Mutlaq Aldhbuiub, Abdulrahman Mosaibah, Abdelrahman Rezk, Abdulrazzaq Alhendi, Mazen~Abo Shal, et~al.
\newblock Masader plus: A new interface for exploring+ 500 arabic nlp datasets.
\newblock {\em arXiv preprint arXiv:2208.00932}, 2022.

\bibitem[\protect\citeauthoryear{Angel \bgroup \em et al.\egroup }{2023}]{angel2023clinical}
Mirana Angel, Anuj Patel, Amal Alachkar, and Pierre Baldi.
\newblock Clinical knowledge and reasoning abilities of large language models in pharmacy: A comparative study on the naplex exam.
\newblock In {\em 2023 Tenth International Conference on Social Networks Analysis, Management and Security (SNAMS)}, pages 1--4. IEEE, 2023.

\bibitem[\protect\citeauthoryear{Bai \bgroup \em et al.\egroup }{2022}]{bai2022training}
Yuntao Bai, Andy Jones, Kamal Ndousse, Amanda Askell, Anna Chen, Nova DasSarma, Dawn Drain, Stanislav Fort, Deep Ganguli, Tom Henighan, et~al.
\newblock Training a helpful and harmless assistant with reinforcement learning from human feedback.
\newblock {\em arXiv preprint arXiv:2204.05862}, 2022.

\bibitem[\protect\citeauthoryear{Beltagy \bgroup \em et al.\egroup }{2019}]{beltagy2019scibert}
Iz~Beltagy, Kyle Lo, and Arman Cohan.
\newblock Scibert: A pretrained language model for scientific text.
\newblock {\em arXiv preprint arXiv:1903.10676}, 2019.

\bibitem[\protect\citeauthoryear{Bender \bgroup \em et al.\egroup }{2021}]{bender2021dangers}
Emily~M Bender, Timnit Gebru, Angelina McMillan-Major, and Shmargaret Shmitchell.
\newblock On the dangers of stochastic parrots: Can language models be too big?
\newblock In {\em Proceedings of the 2021 ACM conference on fairness, accountability, and transparency}, pages 610--623, 2021.

\bibitem[\protect\citeauthoryear{Bengio \bgroup \em et al.\egroup }{2000}]{bengio2000neural}
Yoshua Bengio, R{\'e}jean Ducharme, and Pascal Vincent.
\newblock A neural probabilistic language model.
\newblock {\em Advances in neural information processing systems}, 13, 2000.

\bibitem[\protect\citeauthoryear{Brown \bgroup \em et al.\egroup }{2020}]{brown2020language}
Tom Brown, Benjamin Mann, Nick Ryder, Melanie Subbiah, Jared~D Kaplan, Prafulla Dhariwal, Arvind Neelakantan, Pranav Shyam, Girish Sastry, Amanda Askell, et~al.
\newblock Language models are few-shot learners.
\newblock {\em Advances in neural information processing systems}, 33:1877--1901, 2020.

\bibitem[\protect\citeauthoryear{Chithrananda \bgroup \em et al.\egroup }{2020}]{chithrananda2020chemberta}
Seyone Chithrananda, Gabriel Grand, and Bharath Ramsundar.
\newblock Chemberta: large-scale self-supervised pretraining for molecular property prediction.
\newblock {\em arXiv preprint arXiv:2010.09885}, 2020.

\bibitem[\protect\citeauthoryear{Chowdhery \bgroup \em et al.\egroup }{2023}]{chowdhery2023palm}
Aakanksha Chowdhery, Sharan Narang, Jacob Devlin, Maarten Bosma, Gaurav Mishra, Adam Roberts, Paul Barham, Hyung~Won Chung, Charles Sutton, Sebastian Gehrmann, et~al.
\newblock Palm: Scaling language modeling with pathways.
\newblock {\em Journal of Machine Learning Research}, 24(240):1--113, 2023.

\bibitem[\protect\citeauthoryear{Collobert \bgroup \em et al.\egroup }{2011}]{collobert2011natural}
Ronan Collobert, Jason Weston, L{\'e}on Bottou, Michael Karlen, Koray Kavukcuoglu, and Pavel Kuksa.
\newblock Natural language processing (almost) from scratch.
\newblock {\em Journal of machine learning research}, 12:2493--2537, 2011.

\bibitem[\protect\citeauthoryear{Devlin \bgroup \em et al.\egroup }{2018}]{devlin2018bert}
Jacob Devlin, Ming-Wei Chang, Kenton Lee, and Kristina Toutanova.
\newblock Bert: Pre-training of deep bidirectional transformers for language understanding.
\newblock {\em arXiv preprint arXiv:1810.04805}, 2018.

\bibitem[\protect\citeauthoryear{Graves}{2013}]{graves2013generating}
Alex Graves.
\newblock Generating sequences with recurrent neural networks.
\newblock {\em arXiv preprint arXiv:1308.0850}, 2013.

\bibitem[\protect\citeauthoryear{Hestness \bgroup \em et al.\egroup }{2017}]{hestness2017deep}
Joel Hestness, Sharan Narang, Newsha Ardalani, Gregory Diamos, Heewoo Jun, Hassan Kianinejad, Md~Mostofa~Ali Patwary, Yang Yang, and Yanqi Zhou.
\newblock Deep learning scaling is predictable, empirically.
\newblock {\em arXiv preprint arXiv:1712.00409}, 2017.

\bibitem[\protect\citeauthoryear{Hoffmann \bgroup \em et al.\egroup }{2022}]{hoffmann2022training}
Jordan Hoffmann, Sebastian Borgeaud, Arthur Mensch, Elena Buchatskaya, Trevor Cai, Eliza Rutherford, Diego de~Las Casas, Lisa~Anne Hendricks, Johannes Welbl, Aidan Clark, et~al.
\newblock Training compute-optimal large language models.
\newblock {\em arXiv preprint arXiv:2203.15556}, 2022.

\bibitem[\protect\citeauthoryear{Howard and Ruder}{2018}]{howard2018universal}
Jeremy Howard and Sebastian Ruder.
\newblock Universal language model fine-tuning for text classification.
\newblock {\em arXiv preprint arXiv:1801.06146}, 2018.

\bibitem[\protect\citeauthoryear{Jernite \bgroup \em et al.\egroup }{2022}]{jernite2022data}
Yacine Jernite, Huu Nguyen, Stella Biderman, Anna Rogers, Maraim Masoud, Valentin Danchev, Samson Tan, Alexandra~Sasha Luccioni, Nishant Subramani, Isaac Johnson, et~al.
\newblock Data governance in the age of large-scale data-driven language technology.
\newblock In {\em Proceedings of the 2022 ACM Conference on Fairness, Accountability, and Transparency}, pages 2206--2222, 2022.

\bibitem[\protect\citeauthoryear{Kaplan \bgroup \em et al.\egroup }{2020}]{kaplan2020scaling}
Jared Kaplan, Sam McCandlish, Tom Henighan, Tom~B Brown, Benjamin Chess, Rewon Child, Scott Gray, Alec Radford, Jeffrey Wu, and Dario Amodei.
\newblock Scaling laws for neural language models.
\newblock {\em arXiv preprint arXiv:2001.08361}, 2020.

\bibitem[\protect\citeauthoryear{Karpov \bgroup \em et al.\egroup }{2020}]{karpov2020transformer}
Pavel Karpov, Guillaume Godin, and Igor~V Tetko.
\newblock Transformer-cnn: Swiss knife for qsar modeling and interpretation.
\newblock {\em Journal of cheminformatics}, 12:1--12, 2020.

\bibitem[\protect\citeauthoryear{Kim \bgroup \em et al.\egroup }{2021}]{kim2021changes}
Boseop Kim, HyoungSeok Kim, Sang-Woo Lee, Gichang Lee, Donghyun Kwak, Dong~Hyeon Jeon, Sunghyun Park, Sungju Kim, Seonhoon Kim, Dongpil Seo, et~al.
\newblock What changes can large-scale language models bring? intensive study on hyperclova: Billions-scale korean generative pretrained transformers.
\newblock {\em arXiv preprint arXiv:2109.04650}, 2021.

\bibitem[\protect\citeauthoryear{Kudo and Richardson}{2018}]{kudo2018sentencepiece}
Taku Kudo and John Richardson.
\newblock Sentencepiece: A simple and language independent subword tokenizer and detokenizer for neural text processing.
\newblock {\em arXiv preprint arXiv:1808.06226}, 2018.

\bibitem[\protect\citeauthoryear{Lee \bgroup \em et al.\egroup }{2020}]{lee2020biobert}
Jinhyuk Lee, Wonjin Yoon, Sungdong Kim, Donghyeon Kim, Sunkyu Kim, Chan~Ho So, and Jaewoo Kang.
\newblock Biobert: a pre-trained biomedical language representation model for biomedical text mining.
\newblock {\em Bioinformatics}, 36(4):1234--1240, 2020.

\bibitem[\protect\citeauthoryear{Lewis \bgroup \em et al.\egroup }{2019}]{lewis2019bart}
Mike Lewis, Yinhan Liu, Naman Goyal, Marjan Ghazvininejad, Abdelrahman Mohamed, Omer Levy, Ves Stoyanov, and Luke Zettlemoyer.
\newblock Bart: Denoising sequence-to-sequence pre-training for natural language generation, translation, and comprehension.
\newblock {\em arXiv preprint arXiv:1910.13461}, 2019.

\bibitem[\protect\citeauthoryear{Liu \bgroup \em et al.\egroup }{2019}]{liu2019roberta}
Yinhan Liu, Myle Ott, Naman Goyal, Jingfei Du, Mandar Joshi, Danqi Chen, Omer Levy, Mike Lewis, Luke Zettlemoyer, and Veselin Stoyanov.
\newblock Roberta: A robustly optimized bert pretraining approach.
\newblock {\em arXiv preprint arXiv:1907.11692}, 2019.

\bibitem[\protect\citeauthoryear{Madabushi \bgroup \em et al.\egroup }{2022}]{madabushi2022semeval}
Harish~Tayyar Madabushi, Edward Gow-Smith, Marcos Garcia, Carolina Scarton, Marco Idiart, and Aline Villavicencio.
\newblock Semeval-2022 task 2: Multilingual idiomaticity detection and sentence embedding.
\newblock {\em arXiv preprint arXiv:2204.10050}, 2022.

\bibitem[\protect\citeauthoryear{Mielke \bgroup \em et al.\egroup }{2021}]{mielke2021between}
Sabrina~J Mielke, Zaid Alyafeai, Elizabeth Salesky, Colin Raffel, Manan Dey, Matthias Gall{\'e}, Arun Raja, Chenglei Si, Wilson~Y Lee, Beno{\^\i}t Sagot, et~al.
\newblock Between words and characters: A brief history of open-vocabulary modeling and tokenization in nlp.
\newblock {\em arXiv preprint arXiv:2112.10508}, 2021.

\bibitem[\protect\citeauthoryear{Miikkulainen and Dyer}{1991}]{miikkulainen1991natural}
Risto Miikkulainen and Michael~G Dyer.
\newblock Natural language processing with modular pdp networks and distributed lexicon.
\newblock {\em Cognitive Science}, 15(3):343--399, 1991.

\bibitem[\protect\citeauthoryear{Mikolov \bgroup \em et al.\egroup }{2010}]{mikolov2010recurrent}
Tomas Mikolov, Martin Karafi{\'a}t, Lukas Burget, Jan Cernock{\`y}, and Sanjeev Khudanpur.
\newblock Recurrent neural network based language model.
\newblock In {\em Interspeech}, volume~2, pages 1045--1048. Makuhari, 2010.

\bibitem[\protect\citeauthoryear{Mikolov \bgroup \em et al.\egroup }{2013}]{mikolov2013distributed}
Tomas Mikolov, Ilya Sutskever, Kai Chen, Greg~S Corrado, and Jeff Dean.
\newblock Distributed representations of words and phrases and their compositionality.
\newblock {\em Advances in neural information processing systems}, 26, 2013.

\bibitem[\protect\citeauthoryear{Nguyen \bgroup \em et al.\egroup }{2024}]{nguyen2024hyenadna}
Eric Nguyen, Michael Poli, Marjan Faizi, Armin Thomas, Michael Wornow, Callum Birch-Sykes, Stefano Massaroli, Aman Patel, Clayton Rabideau, Yoshua Bengio, et~al.
\newblock Hyenadna: Long-range genomic sequence modeling at single nucleotide resolution.
\newblock {\em Advances in neural information processing systems}, 36, 2024.

\bibitem[\protect\citeauthoryear{Ouyang \bgroup \em et al.\egroup }{2022}]{ouyang2022training}
Long Ouyang, Jeffrey Wu, Xu~Jiang, Diogo Almeida, Carroll Wainwright, Pamela Mishkin, Chong Zhang, Sandhini Agarwal, Katarina Slama, Alex Ray, et~al.
\newblock Training language models to follow instructions with human feedback.
\newblock {\em Advances in neural information processing systems}, 35:27730--27744, 2022.

\bibitem[\protect\citeauthoryear{Papineni \bgroup \em et al.\egroup }{2002}]{papineni2002bleu}
Kishore Papineni, Salim Roukos, Todd Ward, and Wei-Jing Zhu.
\newblock Bleu: a method for automatic evaluation of machine translation.
\newblock In {\em Proceedings of the 40th annual meeting of the Association for Computational Linguistics}, pages 311--318, 2002.

\bibitem[\protect\citeauthoryear{Pennington \bgroup \em et al.\egroup }{2014}]{pennington2014glove}
Jeffrey Pennington, Richard Socher, and Christopher~D Manning.
\newblock Glove: Global vectors for word representation.
\newblock In {\em Proceedings of the 2014 conference on empirical methods in natural language processing (EMNLP)}, pages 1532--1543, 2014.

\bibitem[\protect\citeauthoryear{Peters \bgroup \em et al.\egroup }{2018}]{peters2018dissecting}
Matthew~E Peters, Mark Neumann, Luke Zettlemoyer, and Wen-tau Yih.
\newblock Dissecting contextual word embeddings: Architecture and representation.
\newblock {\em arXiv preprint arXiv:1808.08949}, 2018.

\bibitem[\protect\citeauthoryear{Radford \bgroup \em et al.\egroup }{2018}]{radford2018improving}
Alec Radford, Karthik Narasimhan, Tim Salimans, Ilya Sutskever, et~al.
\newblock Improving language understanding by generative pre-training.
\newblock 2018.

\bibitem[\protect\citeauthoryear{Radford \bgroup \em et al.\egroup }{2019}]{radford2019language}
Alec Radford, Jeffrey Wu, Rewon Child, David Luan, Dario Amodei, Ilya Sutskever, et~al.
\newblock Language models are unsupervised multitask learners.
\newblock {\em OpenAI blog}, 1(8):9, 2019.

\bibitem[\protect\citeauthoryear{Rae \bgroup \em et al.\egroup }{2021}]{rae2021scaling}
Jack~W Rae, Sebastian Borgeaud, Trevor Cai, Katie Millican, Jordan Hoffmann, Francis Song, John Aslanides, Sarah Henderson, Roman Ring, Susannah Young, et~al.
\newblock Scaling language models: Methods, analysis \& insights from training gopher.
\newblock {\em arXiv preprint arXiv:2112.11446}, 2021.

\bibitem[\protect\citeauthoryear{Raffel \bgroup \em et al.\egroup }{2020}]{raffel2020exploring}
Colin Raffel, Noam Shazeer, Adam Roberts, Katherine Lee, Sharan Narang, Michael Matena, Yanqi Zhou, Wei Li, and Peter~J Liu.
\newblock Exploring the limits of transfer learning with a unified text-to-text transformer.
\newblock {\em Journal of machine learning research}, 21(140):1--67, 2020.

\bibitem[\protect\citeauthoryear{Sanh \bgroup \em et al.\egroup }{2021}]{sanh2021multitask}
Victor Sanh, Albert Webson, Colin Raffel, Stephen~H Bach, Lintang Sutawika, Zaid Alyafeai, Antoine Chaffin, Arnaud Stiegler, Teven~Le Scao, Arun Raja, et~al.
\newblock Multitask prompted training enables zero-shot task generalization.
\newblock {\em arXiv preprint arXiv:2110.08207}, 2021.

\bibitem[\protect\citeauthoryear{Sarzynska-Wawer \bgroup \em et al.\egroup }{2021}]{sarzynska2021detecting}
Justyna Sarzynska-Wawer, Aleksander Wawer, Aleksandra Pawlak, Julia Szymanowska, Izabela Stefaniak, Michal Jarkiewicz, and Lukasz Okruszek.
\newblock Detecting formal thought disorder by deep contextualized word representations.
\newblock {\em Psychiatry Research}, 304:114135, 2021.

\bibitem[\protect\citeauthoryear{Schmidhuber and Heil}{1996}]{schmidhuber1996sequential}
J{\"u}rgen Schmidhuber and Stefan Heil.
\newblock Sequential neural text compression.
\newblock {\em IEEE Transactions on Neural Networks}, 7(1):142--146, 1996.

\bibitem[\protect\citeauthoryear{Schulman \bgroup \em et al.\egroup }{2017}]{schulman2017proximal}
John Schulman, Filip Wolski, Prafulla Dhariwal, Alec Radford, and Oleg Klimov.
\newblock Proximal policy optimization algorithms.
\newblock {\em arXiv preprint arXiv:1707.06347}, 2017.

\bibitem[\protect\citeauthoryear{Segler \bgroup \em et al.\egroup }{2018}]{segler2018planning}
Marwin~HS Segler, Mike Preuss, and Mark~P Waller.
\newblock Planning chemical syntheses with deep neural networks and symbolic ai.
\newblock {\em Nature}, 555(7698):604--610, 2018.

\bibitem[\protect\citeauthoryear{Shahbaz \bgroup \em et al.\egroup }{2019}]{shahbaz2019elmo}
Muhammad Shahbaz, Lalith Suresh, Jennifer Rexford, Nick Feamster, Ori Rottenstreich, and Mukesh Hira.
\newblock Elmo: Source routed multicast for public clouds.
\newblock In {\em Proceedings of the ACM Special Interest Group on Data Communication}, pages 458--471. 2019.

\bibitem[\protect\citeauthoryear{Shibata \bgroup \em et al.\egroup }{1999}]{shibata1999byte}
Yusuxke Shibata, Takuya Kida, Shuichi Fukamachi, Masayuki Takeda, Ayumi Shinohara, Takeshi Shinohara, and Setsuo Arikawa.
\newblock Byte pair encoding: A text compression scheme that accelerates pattern matching.
\newblock 1999.

\bibitem[\protect\citeauthoryear{Shoeybi \bgroup \em et al.\egroup }{2019}]{shoeybi2019megatron}
Mohammad Shoeybi, Mostofa Patwary, Raul Puri, Patrick LeGresley, Jared Casper, and Bryan Catanzaro.
\newblock Megatron-lm: Training multi-billion parameter language models using model parallelism.
\newblock {\em arXiv preprint arXiv:1909.08053}, 2019.

\bibitem[\protect\citeauthoryear{Smith \bgroup \em et al.\egroup }{2022}]{smith2022using}
Shaden Smith, Mostofa Patwary, Brandon Norick, Patrick LeGresley, Samyam Rajbhandari, Jared Casper, Zhun Liu, Shrimai Prabhumoye, George Zerveas, Vijay Korthikanti, et~al.
\newblock Using deepspeed and megatron to train megatron-turing nlg 530b, a large-scale generative language model.
\newblock {\em arXiv preprint arXiv:2201.11990}, 2022.

\bibitem[\protect\citeauthoryear{Sutskever \bgroup \em et al.\egroup }{2011}]{sutskever2011generating}
Ilya Sutskever, James Martens, and Geoffrey~E Hinton.
\newblock Generating text with recurrent neural networks.
\newblock In {\em Proceedings of the 28th international conference on machine learning (ICML-11)}, pages 1017--1024, 2011.

\bibitem[\protect\citeauthoryear{Tay \bgroup \em et al.\egroup }{2022}]{tay2022transcending}
Yi~Tay, Jason Wei, Hyung~Won Chung, Vinh~Q Tran, David~R So, Siamak Shakeri, Xavier Garcia, Huaixiu~Steven Zheng, Jinfeng Rao, Aakanksha Chowdhery, et~al.
\newblock Transcending scaling laws with 0.1\% extra compute.
\newblock {\em arXiv preprint arXiv:2210.11399}, 2022.

\bibitem[\protect\citeauthoryear{Vaswani \bgroup \em et al.\egroup }{2017}]{vaswani2017attention}
Ashish Vaswani, Noam Shazeer, Niki Parmar, Jakob Uszkoreit, Llion Jones, Aidan~N Gomez, {\L}ukasz Kaiser, and Illia Polosukhin.
\newblock Attention is all you need.
\newblock {\em Advances in neural information processing systems}, 30, 2017.

\bibitem[\protect\citeauthoryear{Vinyals and Le}{2015}]{vinyals2015neural}
Oriol Vinyals and Quoc Le.
\newblock A neural conversational model.
\newblock {\em arXiv preprint arXiv:1506.05869}, 2015.

\bibitem[\protect\citeauthoryear{Wang \bgroup \em et al.\egroup }{2021}]{wang2021ernie}
Shuohuan Wang, Yu~Sun, Yang Xiang, Zhihua Wu, Siyu Ding, Weibao Gong, Shikun Feng, Junyuan Shang, Yanbin Zhao, Chao Pang, et~al.
\newblock Ernie 3.0 titan: Exploring larger-scale knowledge enhanced pre-training for language understanding and generation.
\newblock {\em arXiv preprint arXiv:2112.12731}, 2021.

\bibitem[\protect\citeauthoryear{Wang \bgroup \em et al.\egroup }{2023a}]{wang2023openchat}
Guan Wang, Sijie Cheng, Xianyuan Zhan, Xiangang Li, Sen Song, and Yang Liu.
\newblock Openchat: Advancing open-source language models with mixed-quality data.
\newblock {\em arXiv preprint arXiv:2309.11235}, 2023.

\bibitem[\protect\citeauthoryear{Wang \bgroup \em et al.\egroup }{2023b}]{wang2023uni}
Xi~Wang, Ruichu Gu, Zhiyuan Chen, Yongge Li, Xiaohong Ji, Guolin Ke, and Han Wen.
\newblock Uni-rna: universal pre-trained models revolutionize rna research.
\newblock {\em bioRxiv}, pages 2023--07, 2023.

\bibitem[\protect\citeauthoryear{Wei \bgroup \em et al.\egroup }{2022}]{wei2022emergent}
Jason Wei, Yi~Tay, Rishi Bommasani, Colin Raffel, Barret Zoph, Sebastian Borgeaud, Dani Yogatama, Maarten Bosma, Denny Zhou, Donald Metzler, et~al.
\newblock Emergent abilities of large language models.
\newblock {\em arXiv preprint arXiv:2206.07682}, 2022.

\bibitem[\protect\citeauthoryear{Yang \bgroup \em et al.\egroup }{2019}]{yang2019xlnet}
Zhilin Yang, Zihang Dai, Yiming Yang, Jaime Carbonell, Russ~R Salakhutdinov, and Quoc~V Le.
\newblock Xlnet: Generalized autoregressive pretraining for language understanding.
\newblock {\em Advances in neural information processing systems}, 32, 2019.

\bibitem[\protect\citeauthoryear{Zeng \bgroup \em et al.\egroup }{2016}]{zeng2016convolutional}
Haoyang Zeng, Matthew~D Edwards, Ge~Liu, and David~K Gifford.
\newblock Convolutional neural network architectures for predicting dna--protein binding.
\newblock {\em Bioinformatics}, 32(12):i121--i127, 2016.

\bibitem[\protect\citeauthoryear{Zeng \bgroup \em et al.\egroup }{2021}]{zeng2021pangu}
Wei Zeng, Xiaozhe Ren, Teng Su, Hui Wang, Yi~Liao, Zhiwei Wang, Xin Jiang, ZhenZhang Yang, Kaisheng Wang, Xiaoda Zhang, et~al.
\newblock Pangu-$alpha $: Large-scale autoregressive pretrained chinese language models with auto-parallel computation.
\newblock {\em arXiv preprint arXiv:2104.12369}, 2021.

\bibitem[\protect\citeauthoryear{Zeng \bgroup \em et al.\egroup }{2022}]{zeng2022glm}
Aohan Zeng, Xiao Liu, Zhengxiao Du, Zihan Wang, Hanyu Lai, Ming Ding, Zhuoyi Yang, Yifan Xu, Wendi Zheng, Xiao Xia, et~al.
\newblock Glm-130b: An open bilingual pre-trained model.
\newblock {\em arXiv preprint arXiv:2210.02414}, 2022.

\bibitem[\protect\citeauthoryear{Zhang \bgroup \em et al.\egroup }{2019}]{zhang2019ernie}
Zhengyan Zhang, Xu~Han, Zhiyuan Liu, Xin Jiang, Maosong Sun, and Qun Liu.
\newblock Ernie: Enhanced language representation with informative entities.
\newblock {\em arXiv preprint arXiv:1905.07129}, 2019.

\bibitem[\protect\citeauthoryear{Zhang \bgroup \em et al.\egroup }{2022}]{zhang2022opt}
Susan Zhang, Stephen Roller, Naman Goyal, Mikel Artetxe, Moya Chen, Shuohui Chen, Christopher Dewan, Mona Diab, Xian Li, Xi~Victoria Lin, et~al.
\newblock Opt: Open pre-trained transformer language models.
\newblock {\em arXiv preprint arXiv:2205.01068}, 2022.

\bibitem[\protect\citeauthoryear{Zhang \bgroup \em et al.\egroup }{2023}]{zhang2023dnagpt}
Daoan Zhang, Weitong Zhang, Bing He, Jianguo Zhang, Chenchen Qin, and Jianhua Yao.
\newblock Dnagpt: A generalized pretrained tool for multiple dna sequence analysis tasks.
\newblock {\em bioRxiv}, pages 2023--07, 2023.

\bibitem[\protect\citeauthoryear{Zhao \bgroup \em et al.\egroup }{2020}]{zhao2020limitations}
Wei Zhao, Goran Glava{\v{s}}, Maxime Peyrard, Yang Gao, Robert West, and Steffen Eger.
\newblock On the limitations of cross-lingual encoders as exposed by reference-free machine translation evaluation.
\newblock {\em arXiv preprint arXiv:2005.01196}, 2020.

\end{thebibliography}

\clearpage
\newpage
\onecolumn
\appendix
\section*{Appendix}

\ignore{\begin{figure*}[!h]
\begin{center}
\includegraphics[width=6.0in]{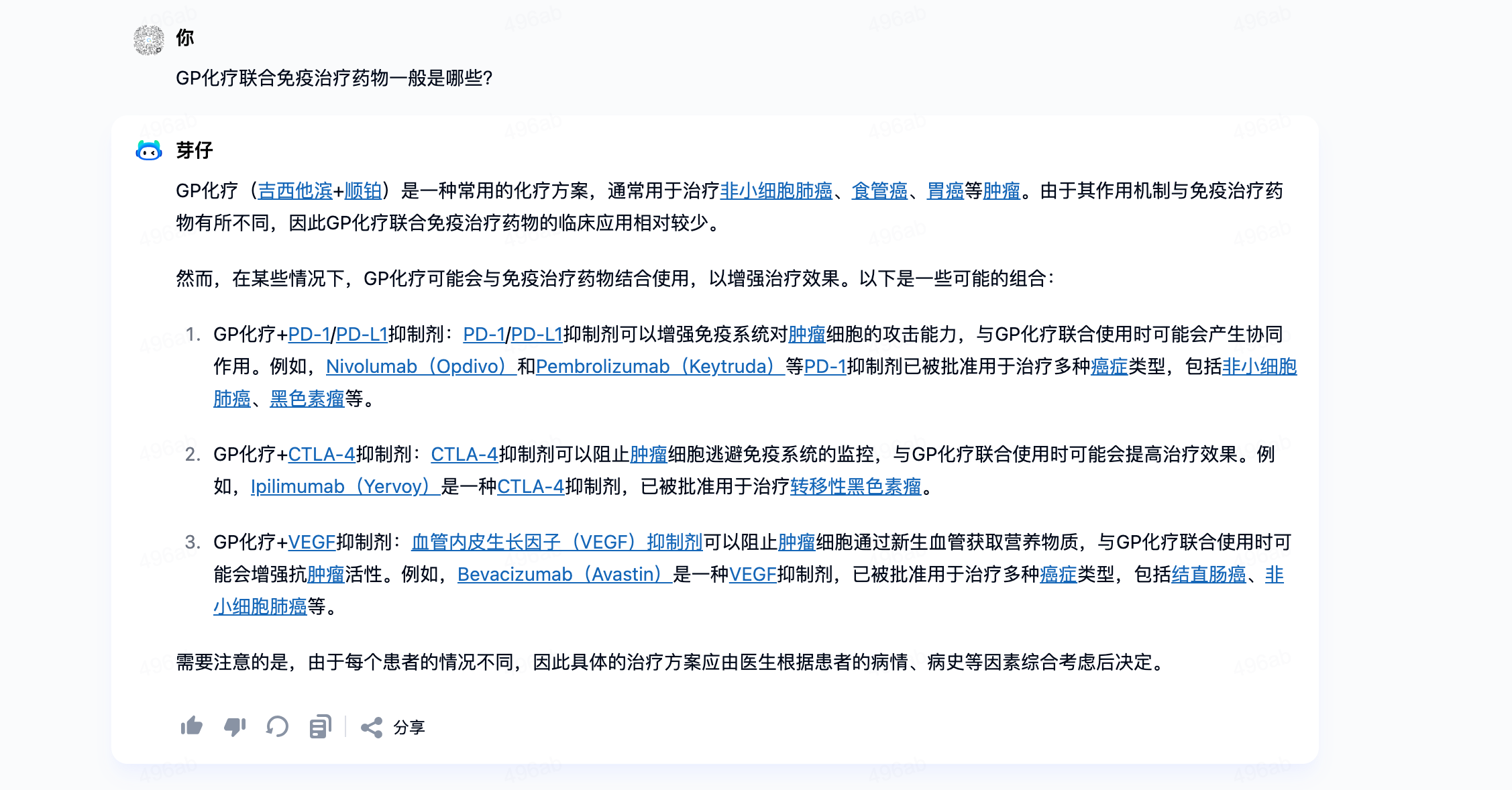}
\end{center}
\caption{Case study.} 
\label{fig:NO.1}
\end{figure*}
}

\subsection*{A. Case Study}

\begin{CJK}{UTF8}{gbsn}
\begin{table*}[!ht]
    \centering
    \small
    \resizebox{\linewidth}{!}{
    \begin{tabular}{c|l}
        \cline{1-2}
        \hline
        Language  & PharmaGPT0.7 \\
        \hline
        ZH & 用户：\\
        &GP化疗联合免疫治疗药物一般是哪些?\\
        &PharmaGPT:\\
        &GP化疗(吉西他滨+顺铂)是一种常用的化疗方案,通常用于治疗非小细胞肺癌、食管癌、胃癌等肿瘤。由于其作用机制与免疫治疗药物有所不同,因此GP化疗联合免疫治疗药物的临床应用相对较少。\\
        &然而,在某些情况下,GP化疗可能会与免疫治疗药物结合使用,以增强治疗效果。以下是一些可能的组合:\\
        &1. GP化疗+PD-1/PD-L1抑制剂:PD-1/PD-L1抑制剂可以增强免疫系统对肿瘤细胞的攻击能力。与GP化疗联合使用时可能会产生协同作用,例如, Nivolumab (Opdivo) 和Pembrolizumab (Keytruda) \\
        &等PD-1抑制剂已被批准用于治疗多种癌症类型,包括非小细胞肺癌、黑色素瘤等。\\ 
        &2. GP化疗+CTLA-4抑制剂: CTLA-4抑制剂可以阻止肿瘤细胞逃避免疫系统的监控,与GP化疗联合使用时可能会提高治疗效果。例如, Ipilimumab (Yervoy) 是一种CTLA-4抑制剂,已被批准用于治\\
        &疗转移性黑色素瘤。\\
        &3. GP化疗+VEGF抑制剂:血管内皮生长因子(VEGF)抑制剂可以阻止肿瘤细胞通过新生血管获取营养物质,与GP化疗联合使用时可能会增强抗肿瘤活性。例如, Bevacizumab (Avastin) 是一种VEGF\\
        &抑制剂,已被批准用于治疗多种癌症类型,包括结直肠癌、非小细胞肺癌等。\\
        &需要注意的是,由于每个患者的情况不同,因此具体的治疗方案应由医生根据患者的病情、病史等因素综合考虑后决定。 \\
        EN  & User: \\
&What are the common immunotherapy drugs combined with GP chemotherapy?\\
&PharmaGPT:\\
&GP chemotherapy (gemcitabine + cisplatin) is a commonly used chemotherapy regimen, usually for treating non-small cell lung cancer, esophageal cancer, gastric cancer, and other tumors. Due to its different\\
&mechanism of action compared to immunotherapy drugs, the clinical application of GP chemotherapy combined with immunotherapy drugs is relatively rare.\\
&However, in some cases, GP chemotherapy may be combined with immunotherapy drugs to enhance the treatment effect. The following are some possible combinations:\\
&1. GP chemotherapy + PD-1/PD-L1 inhibitors: PD-1/PD-L1 inhibitors can enhance the immune system's ability to attack tumor cells. When used in combination with GP chemotherapy, they may produce\\
&a synergistic effect. For example, Nivolumab (Opdivo) and Pembrolizumab (Keytruda) \\
&and other PD-1 inhibitors have been approved for the treatment of various cancer types, including non-small cell lung cancer, melanoma, etc.\\
&2. GP chemotherapy + CTLA-4 inhibitors: CTLA-4 inhibitors can prevent tumor cells from escaping immune system surveillance. When used in combination with GP chemotherapy, they may improve the\\
&treatment effect. For example, Ipilimumab (Yervoy) is a CTLA-4 inhibitor that has been approved for the treatment of\\
&metastatic melanoma.\\
&3. GP chemotherapy + VEGF inhibitors: Vascular endothelial growth factor (VEGF) inhibitors can prevent tumor cells from obtaining nutrients through neovascularization. When used in combination with\\
&GP chemotherapy, they may enhance anti-tumor activity. For example, Bevacizumab (Avastin) is a VEGF\\
&inhibitor that has been approved for the treatment of various cancer types, including colorectal cancer, non-small cell lung cancer, etc.\\
&It should be noted that due to the different conditions of each patient, the specific treatment plan should be determined by the doctor after comprehensive consideration of the patient's condition, medical history, and other factors. \\
        \hline






        \cline{1-2}
    \end{tabular}
    }
    \caption{Case study.}
    \label{tab:case-study}
\end{table*}
\end{CJK}

\ignore{In this section, we present several compelling examples demonstrating the superior specialization of PharmaGPT in the fields of biomedicine and chemistry, surpassing that of other models. These instances underscore the model's remarkable ability to navigate and interpret complex domain-specific information, setting a new benchmark for performance in this space.

The response given by PharmaGPT provides a comprehensive overview of the combination of GP chemotherapy (gemcitabine plus cisplatin) with immunotherapy drugs, which is a topic of significant interest in the field of oncology. This combination strategy is often utilized in the treatment of various cancers, including non-small cell lung cancer, esophageal cancer, and gastric cancer. The explanation why this approach is considered beneficial, despite its relatively rare clinical application, involves a detailed understanding of the distinct mechanisms of action between chemotherapy and immunotherapy agents.

Chemotherapy, such as the GP regimen, primarily targets rapidly dividing cells, including cancer cells, by causing DNA damage or interfering with cell division. On the other hand, immunotherapy aims to enhance the body's immune system to recognize and fight cancer cells more effectively. The combination of these therapies is thought to potentially create a synergistic effect, whereby chemotherapy not only reduces the tumor burden but also may alter the tumor microenvironment in a way that makes cancer cells more susceptible to immune attack.

The specific combinations mentioned, including GP chemotherapy with PD-1/PD-L1 inhibitors (e.g., Nivolumab, Pembrolizumab), CTLA-4 inhibitors (e.g., Ipilimumab), and VEGF inhibitors (e.g., Bevacizumab), represent strategic attempts to target different pathways involved in cancer progression and immune evasion. Each of these combinations is designed to enhance the overall anti-tumor activity by not only directly killing tumor cells but also by preventing the tumor from evading immune surveillance and by inhibiting the tumor's ability to sustain itself through angiogenesis.}

In this section, we detail specific examples that demonstrate the superior functionality of PharmaGPT0.7, particularly in the fields of biomedicine and oncological treatments. These instances highlight the model's advanced capabilities in handling and elucidating complex, domain-specific information, which sets a new industry standard for AI-driven data interpretation in biomedical sciences.

\textbf{Detailed Response Analysis of PharmaGPT0.7}:
The responses provided by PharmaGPT0.7 for both Chinese (ZH) and English (EN) queries about the use of GP chemotherapy in combination with immunotherapy illustrate its exceptional proficiency. GP chemotherapy, consisting of gemcitabine and cisplatin, is a regimen noted for its effectiveness against various tumors such as non-small cell lung cancer, esophageal cancer, and gastric cancer.

\textbf{Specialization and Depth}:
PharmaGPT0.7 not only identifies the typical combinations of GP chemotherapy with immunotherapy drugs but also explains the synergy between these treatments in a clinically relevant context:
\begin{itemize}
\item \textbf{PD-1/PD-L1 Inhibitors}: The model correctly identifies drugs like Nivolumab and Pembrolizumab, explaining their mechanism to enhance immune system capability against tumor cells, a crucial piece of information for oncologists.
\item \textbf{CTLA-4 Inhibitors}: Detailing the role of CTLA-4 inhibitors like Ipilimumab in preventing tumor cells from evading immune surveillance, PharmaGPT0.7 provides actionable insights into their combination with chemotherapy.
\item \textbf{VEGF Inhibitors}: By mentioning Bevacizumab and its role in inhibiting neovascularization, the model showcases a deep understanding of how angiogenesis inhibitors can complement cytotoxic chemotherapy.
\end{itemize}

\textbf{Clinical Relevance}:
PharmaGPT0.7 further enhances its responses by noting that the choice of specific immunotherapy drugs alongside GP chemotherapy must be personalized, taking into consideration the patient's medical history and current health condition. This approach reflects a sophisticated understanding of oncology, where treatment customization is paramount.

\textbf{Comparison with Other Models}:
Unlike other models, PharmaGPT0.7’s responses are not only rich in content but are also specifically tailored to reflect the latest trends and clinical trials in oncology. This allows healthcare professionals to receive the most current and relevant information, aiding in better decision-making processes.

\textbf{Conclusion}:
PharmaGPT0.7’s detailed, accurate, and contextually relevant responses set it apart in the landscape of AI-driven tools in biomedicine, making it an invaluable resource for professionals who require up-to-date and detailed information to make informed clinical decisions.

\ignore{
\subsection*{B. Dynamic Vertical Domain Data}

\begin{figure}[!h]
\begin{center}
\includegraphics[width=2.0in]{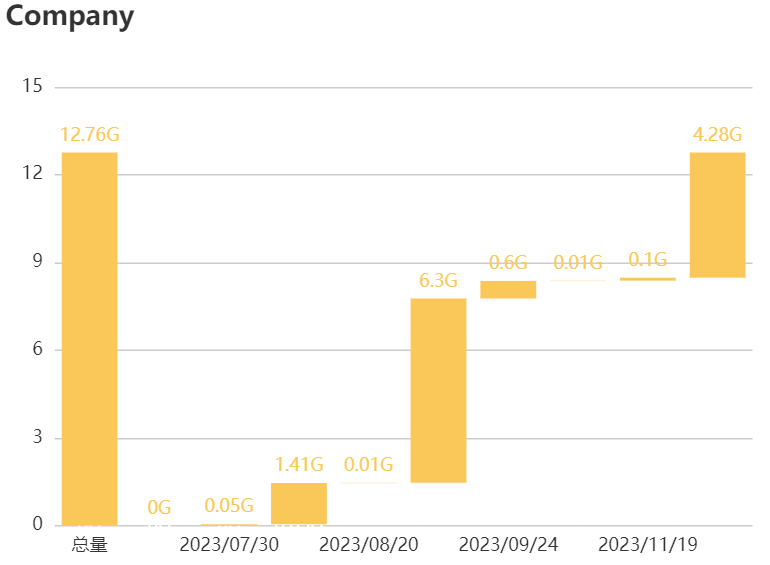}
\end{center}
\caption{Dynamically growing dataset of Bio-Pharmaceutical company.} 
\label{fig:company}
\end{figure}

\begin{figure}[!h]
\begin{center}
\includegraphics[width=2.0in]{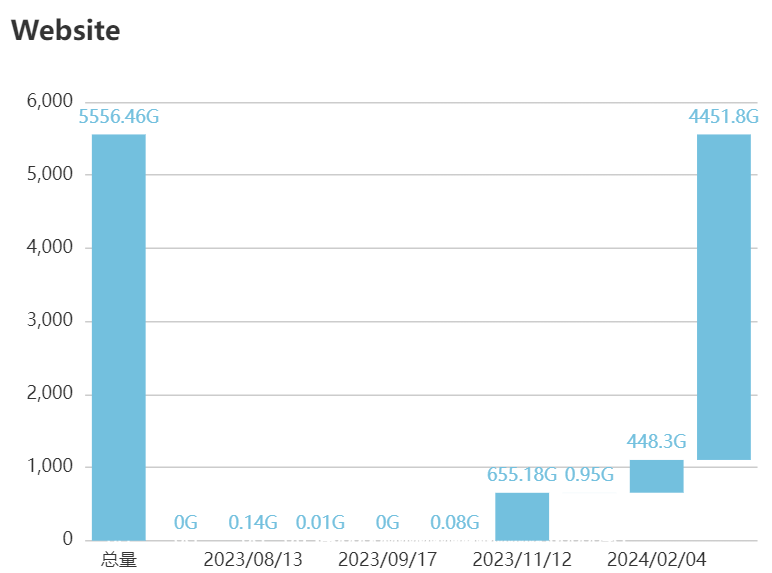}
\end{center}
\caption{Dynamically growing dataset of website.} 
\label{fig:website}
\end{figure}

\begin{figure}[!h]
\begin{center}
\includegraphics[width=2.6in]{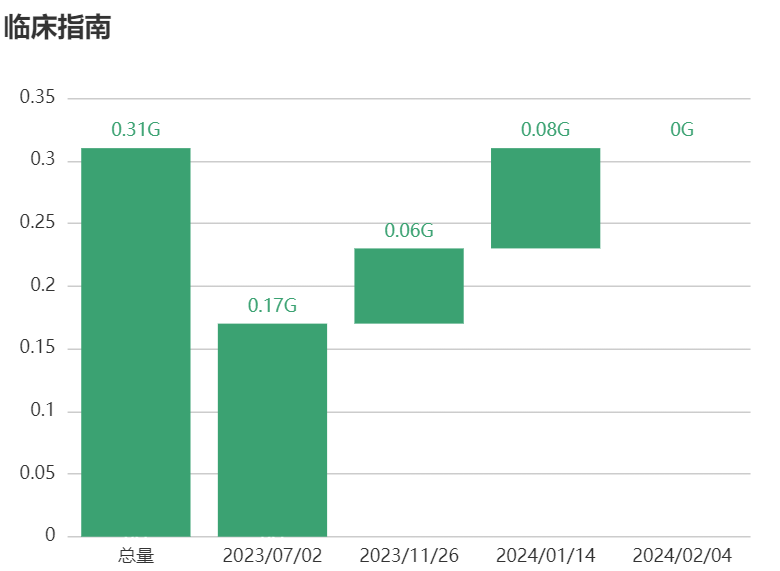}
\end{center}
\caption{Dynamically growing dataset of clinical reports.} 
\label{fig:clinic}
\end{figure}

\begin{figure}[!h]
\begin{center}
\includegraphics[width=2.5in]{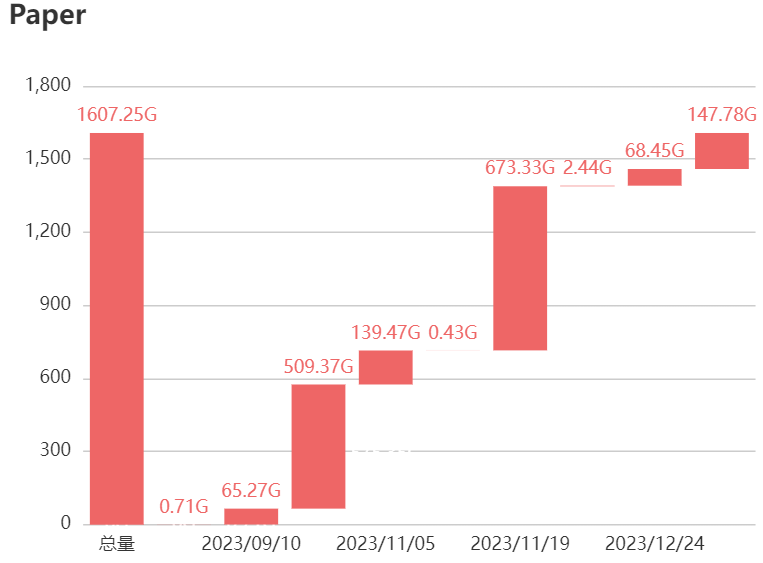}
\end{center}
\caption{Dynamically growing dataset of academic papers.} 
\label{fig:paper}
\end{figure}

}

\begin{figure*}[!t]
\begin{center}
\includegraphics[width=6.0in]{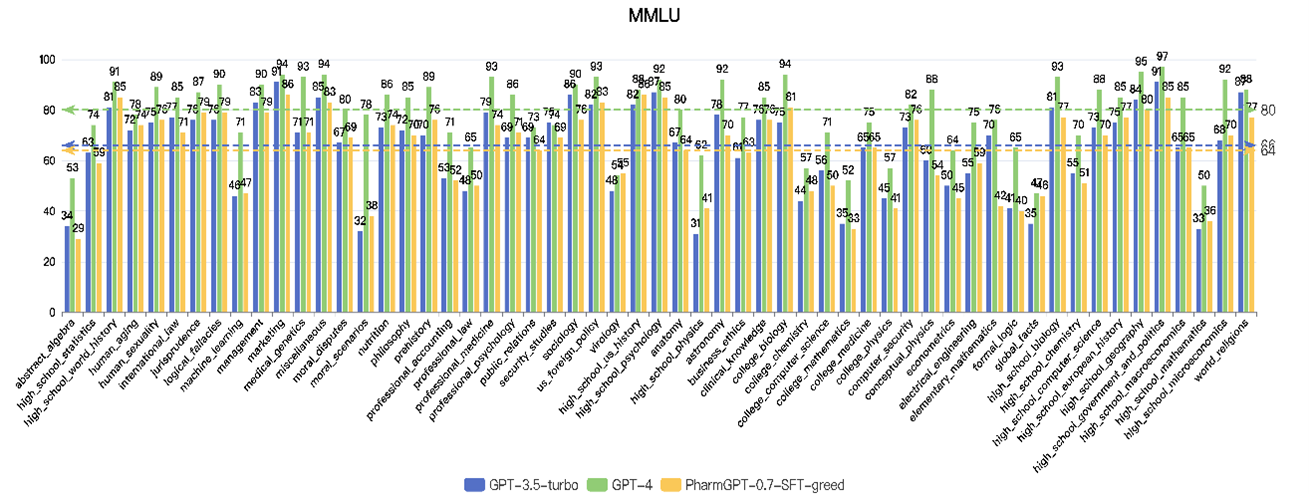}
\end{center}
\caption{Test Performance of PharmaGPT on the General Test Set MMLU.} 
\label{fig:mmlu}
\end{figure*}

\subsection*{C. Comparison on the MMLU Dataset}
We conducted a comparative analysis of our model, PharmaGPT, with leading Large Language Models (LLMs) such as GPT-4, using the widely recognized MMLU dataset. Our findings are twofold: firstly, PharmaGPT achieves impressive results in general capabilities despite having significantly fewer parameters than its counterparts. Secondly, it demonstrates superior proficiency in specialized domains including biomedicine, chemistry, and related fields, surpassing the advanced capabilities of general-purpose large models.

\textbf{Overall Strong Performance}
PharmaGPT consistently maintains MMLU scores in the 80-90\% range across nearly all tasks, illustrating its robust capabilities in the biomedical and life sciences domains. This high performance indicates effective fine-tuning on a substantial corpus of relevant scientific literature.

\textbf{Outperforming GPT-3.5-turbo}
In the majority of tasks, particularly those related to biology, medicine, anatomy, and physiology, PharmaGPT not only achieves higher scores than GPT-3.5-turbo but often by significant margins. This highlights the benefits of domain-specific training, which enhances PharmaGPT's performance in biomedical question answering compared to more generalized models like GPT-3.5.

\textbf{Comparable to GPT-4}
Notably, PharmaGPT scores very closely to GPT-4 on a broad spectrum of topics and slightly outperforms GPT-4 in areas such as physiology, health sciences, and biology. This achievement is particularly impressive considering GPT-4's larger scale, indicating that PharmaGPT’s focused training regimen allows it to match, and occasionally surpass, GPT-4’s biomedical knowledge without the need for vast computational resources.

\textbf{Areas for Improvement}
Despite its strengths, PharmaGPT shows room for improvement in topics such as world religions, philosophy, and other miscellaneous areas, where it significantly lags behind GPT-4. This limitation likely reflects the scope and nature of PharmaGPT's training data. However, given the model's specialized focus, this minor underperformance in non-core subjects is understandable and does not detract from its overall effectiveness for biomedical applications.

The results underscore PharmaGPT's state-of-the-art performance in biomedical language understanding, positioning it as a powerful tool for domain-specific NLP tasks. Its focused training approach not only allows it to surpass GPT-3.5 but also to match, and in some cases exceed, GPT-4's capabilities in key life science and healthcare topics, all while operating on a smaller scale. This makes PharmaGPT an attractive option for biomedical researchers and organizations seeking high accuracy and efficiency in their NLP implementations.

\end{document}